\documentclass{article} 
\usepackage{iclr2023_conference,times}

\usepackage{amsmath,amsfonts,bm}

\def\eqref#1{equation~\ref{#1}}

\def\1{\bm{1}}

\DeclareMathAlphabet{\mathsfit}{\encodingdefault}{\sfdefault}{m}{sl}
\SetMathAlphabet{\mathsfit}{bold}{\encodingdefault}{\sfdefault}{bx}{n}

\usepackage{hyperref}
\usepackage{url}
\usepackage{colortbl}
\usepackage{amsfonts}
\usepackage{graphicx}
\usepackage{subcaption}
\usepackage{mathrsfs}
\usepackage{wrapfig}
\usepackage{multirow}

\usepackage{algorithm}
\usepackage{algpseudocode}
\newcommand{\STATEnonum}{\item[]}

\makeatletter
\renewcommand\paragraph{\@startsection{paragraph}{4}{\z@}%
            {-2.5ex\@plus -1ex \@minus -.25ex}%
            {1.25ex \@plus .25ex}%
            {\normalfont\normalsize\bfseries}}
\makeatother
\usepackage{enumitem}

\title{Synthetic Health-related\\
Longitudinal Data with Mixed-type Variables\\
Generated using Diffusion Models}

\author{Nicholas I-Hsien Kuo, Louisa Jorm, Sebastiano Barbieri\\
Centre for Big Data Research in Health (CBDRH)\\
The University of New South Wales, Sydney, Australia\\
\footnotesize{\textcolor{white}{*}}\\
Corresponding author: Nicholas I-Hsien Kuo (\texttt{n.kuo@unsw.edu.au})
}

\iclrfinalcopy 
\begin{document}

\maketitle

\begin{abstract}
This paper presents a novel approach to simulating electronic health records (EHRs) using diffusion probabilistic models (DPMs). Specifically, we demonstrate the effectiveness of DPMs in synthesising longitudinal EHRs that capture mixed-type variables, including numeric, binary, and categorical variables. To our knowledge, this represents the first use of DPMs for this purpose. We compared our DPM-simulated datasets to previous state-of-the-art results based on generative adversarial networks (GANs) for two clinical applications: acute hypotension and human immunodeficiency virus (ART for HIV). Given the lack of similar previous studies in DPMs, a core component of our work involves exploring the advantages and caveats of employing DPMs across a wide range of aspects. In addition to assessing the realism of the synthetic datasets, we also trained reinforcement learning (RL) agents on the synthetic data to evaluate their utility for supporting the development of downstream machine learning models. Finally, we estimated that our DPM-simulated datasets are secure and posed a low patient exposure risk for public access.\\

\hspace*{17.5mm}Generative Adversarial Networks, Diffusion Porbabilistic Models,\\
\hspace*{17.5mm}Hypotension, ART for HIV 
\end{abstract}

\section*{Ethics Statement}
This study was approved by the University of New South Wales' human research ethics committee (application HC210661). We based our synthetic acute hypotension dataset on MIMIC-III~\citep{johnson2016mimic}, and our synthetic HIV dataset on EuResist~\citep{zazzi2012predicting}. For patients in MIMIC-III requirement for individual consent was waived because the project did not impact clinical care and all protected health information was deidentified~\citep{johnson2016mimic}. For people in the EuResist integrated database, all data providers obtained informed consent for the execution of retrospective studies and inclusion in merged cohorts \citep{prosperi2010antiretroviral}.

\section{Introduction}
Healthcare data is a crucial resource for the advancement of machine learning (ML) algorithms, including the field of \textit{reinforcement learning} (RL)~\citep{sutton2018reinforcement}. RL is trained to learn an optimal behaviour policy to match actions (\textit{i.e.,} treatment selections) to the environment (\textit{i.e.,} a patient's clinical state); and it has the potential to significantly improve healthcare~\citep{komorowski2018artificial}. However, privacy regulations (see \citet{nosowsky2006health}, \citet{okeefe2010privacy}, and \citet{bentzen2021remove}) restrict the use of health-related data, limiting the availability of real-world datasets for research. This scarcity of data negatively impacts the development of ML algorithms, as they require large, diversified datasets to effectively learn and improve. Synthetic data, generated through the use of generative models, offers a solution to this challenge~\citep{kuo2022health} . By creating highly realistic synthetic datasets, the research community can develop, test, and compare ML algorithms in a controlled environment, without compromising privacy.

\begin{figure}[t]
    \centering
    \includegraphics[width = 0.85\linewidth]{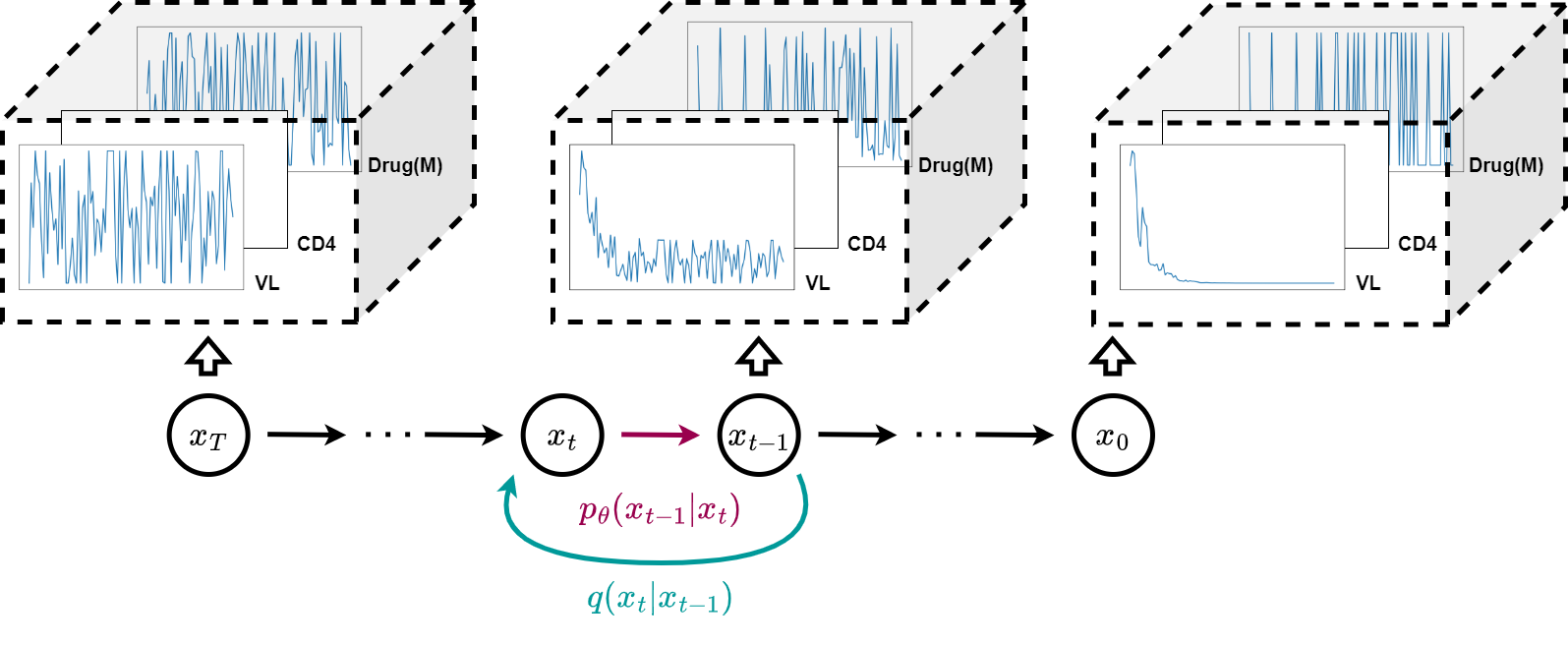}
    \caption{\label{Fig:DPMFrame}Using the DPM framework to generate synthetic sequential data.\\
    The DPM framework consists a forward diffusion process (in cyan) and a reverse diffusion process (in magenta) to process sequential data $x_t$. The goal is to denoise the data at each timestep iteratively, resulting in a set of clean and novel time-series variables. Refer to Section \ref{Sec:DPMFramework} for details on the DPM.
    }
\end{figure}

\textit{Generative Adversarial Networks} (GANs)~\citep{goodfellow2014generative, arjovsky2017wasserstein, gulrajani2017improved} have proven to be effective generative models, with applications found on images~\citep{yu2018generative}, texts~\citep{xu2018diversity}, and audios~\citep{pascual2017segan}. However, training GAN-based models is notoriously difficult because they suffer from \textit{unstable training}~\citep{thanh2019improving} and \textit{mode collapse}~\citep{goodfellow2016nips}. The former decreases the quality of the synthetic data and the latter reduces diversity. Both phenomena cause GANs to generate ill-represented patient states; and this would negatively impact the utility of the synthetic dataset, causing downstream machine learning models to learn biases that may inflict patient harm~\citep{challen2019artificial}.

Lately, the \textit{diffusion probabilistic models} (DPMs)~\citep{sohl2015deep, ho2020denoising} shown in Figure \ref{Fig:DPMFrame} are emerging as one of the top alternative generative frameworks. Similar to GANs, DPMs are extensively applied in images~\citep{dhariwal2021diffusion}, texts~\citep{austin2021structured}, and audios~\citep{kong2020diffwave}. In addition, \citet{ramesh2022hierarchical} demonstrated that their model (\textit{i.e.,} DELL-E2) could generate highly flexible and creative images with DPMs from text prompts. Using the Fr\'echet inception distance (FID)~\citep{heusel2017gans}, \citeauthor{ramesh2022hierarchical} further demonstrated that DPMs achieved higher realism than various GAN-based models. \citet{dhariwal2021diffusion} reported similar findings that DPMs beat GANs on image synthesis.

More recently, DPMs are starting to find applications for \textit{electronic health records} (EHRs). In a study conducted by \citet{zheng2022diffusion}, the authors demonstrated that DPMs can be utilised for the imputation of missing values in clinical tabular data. However to support the developmet of RL, DPMs would need to overcome unique challenges to generate synthetic time-series data over mixed-type variables. To address this, we propose a novel DPM (see Section \ref{Sec:Methods}) that is capable of effectively generating time-series clinical datasets for acute hypotension, sepsis, and the antiretroviral therapy for human immunodeficiency virus (ART for HIV) (see Section \ref{Sec:TruthData}). This constitutes a technical contribution to the field.

In this study, we aim to produce synthetic datasets that can be publicly distributed to hasten research in machine learning. This work marks the first use of DPM to generate synthetic longitudinal EHRs that capture mixed-type variables. As such, a crucial component of our work involves investigating the advantages and caveats of utilising DPMs for this purpose. To assess the validity of our simulated datasets, we conducted three distinct evaluations: (1) a comparison of the statistical properties of synthetic variables with real data (refer to Sections \ref{Sec:IndReal} -- \ref{Sec:Diversity}); (2) a comparison of the \textit{utility}~\citep{rankin2020reliability} of the datasets for developing RL agents (refer to Section \ref{Sec:UtilityInves}); and (3) an estimation of the \textit{patient disclosure risk}~\citep{el2020evaluating} (refer to Section \ref{Sec:SecEst}).\footnote{To facilitate future research, our code and synthetic dataset will be made available after the paper is published. Follow our research progress on \url{https://healthgym.ai/}.}

\newpage
\section{Related Work}\label{Sec:RelatedWork}

This section discusses training difficulties in GANs and some existing clinical applications of GANs.

\subsection{The Difficulty of Fine-tuning GANs}
GANs consist of two sub-networks -- a generator and a discriminator\footnote{WGAN~\citep{arjovsky2017wasserstein} replaces the discriminator with a critic to rate the realisticness of all inputs.}. These sub-networks participate in a two-player minimax game where the goal is to find a Nash equilibrium~\citep{goodfellow2014generative}. The generator creates synthetic data from a random latent prior, while the discriminator aims to determine the authenticity of the generated data by comparing it to real data. Ideally, when the discriminator has optimised and the difference between real and generated data is minimized, the generator has learned to model the underlying probability distribution of the real data.

The training of GANs interleaves the updates of the two sub-networks. This interplay can result in instability in the training process~\citep{kodali2017convergence, mescheder2018training}, which causes fluctuations in the loss over time~\citep{thanh2019improving}, as well as \textit{mode collapse}~\citep{goodfellow2016nips}, where the generator outputs the same sample repeatedly, reducing diversity in the generated data. To mitigate these challenges, prior studies have proposed several implementations to improve the stability and convergence of GAN training, including modifications to the network architecture~\citep{radford2015unsupervised}, learning objectives~\citep{arjovsky2017wasserstein, gulrajani2017improved}, and auxiliary experimental setups~\citep{salimans2016improved, sonderby2016amortised}. Additionally, researchers have proposed techniques to mitigate mode collapse by measuring the diversity of the generated data through features learned within the GAN sub-networks. This has been achieved through methods such as \textit{minibatch discrimination}~\citep{salimans2016improved} and moment matching~\citep{li2017mmd}.

There are also many theoretical work on GANs that emphasises the optimality on the minima~\citep{nagarajan2017gradient, mescheder2017numerics, mescheder2018training} as well as the convexity of the learning objectives~\citep{kodali2017convergence}. One approach enforces Lipschitz constraints on the discriminator network~\citep{gulrajani2017improved, liu2019spectral}; and another line of research has focused on improving the design of the discriminator or using multiple discriminators~\citep{srivastava2017veegan, mordido2020microbatchgan, thanh2020catastrophic} to reduce the risk of \textit{catastrophic forgetting}~\citep{mccloskey1989catastrophic, kuo2021AAAI} -- the abrupt lost of learnt knowledge by repetitive incremental update -- in the discriminator to mitigate mode collapse.

\subsection{GANs in the Healthcare Domain}
While GANs have made significant advancements over the years, they still face certain limitations in the medical domain. Despite extensive research, GANs have primarily been limited to synthesising one type of data. For example, \citet{choi2017generating} (\textit{i.e.,} MedGAN), \citet{xie2018differentially}, and \citet{camino2018generating} only support the generation of discrete variables; while \citet{beaulieu2019privacy} only support numeric variable simulations. Despite recent progress, GAN-generated datasets are largely static in nature~\citep{park2018data, lu2019empirical, yoon2020anonymization, walia2020synthesising} and may only be suitable for developing predictive algorithms. There is a scarcity of literature on GAN-generated datasets suitable for RL agents~\citep{li2021generating, kuo2022health}, and hence there remains a need for further research in this area to address these limitations.

MedGAN~\citep{choi2017generating} has gained popularity in clinical research and is frequently used as a baseline model~\citep{baowaly2019synthesizing, baowaly2019realistic, torfi2020corgan}. Despite its popularity, a study by \citet{goncalves2020generation} found that MedGAN failed to accurately represent multivariate categorical medical data as it performed unfavourably on the log-cluster~\citep{woo2009global} metric.

Similarly, the Health Gym GAN~\citep{kuo2022health}, capable of generating mixed-type time-series clinical data, was found to be susceptible to mode collapse. In an extended study by the same panel of authors~\citep{kuo2022generating}, they found that mode collapse negatively impacted the utility of the synthetic dataset because patients of minority ethnicity could be neglected during the synthesis procedure. To mitigate this issue, the authors stored features from the real data in an external buffer during training and replay them to the generator during synthesis. \citet{marchesi2022mitigating} also found that by using a conditional architecture, the Health Gym GAN could synthesise data that captured a higher level of detail in the real data feature space.

In conclusion, applications of GANs in the medical domain are limited by mode collapse and training instability. Moreover, \citet{kuo2022generating} reported that previous solutions for preventing mode collapse in computer vision~\citep{larsen2016autoencoding, salimans2016improved, li2017mmd, mangalam2021overcoming} prove ineffective for mixed-type time-series clinical data. As an alternative, DPMs may circumvent these limitations, as they are not known for mode collapse and training instability.

\section{Diffusion Probabilistic Models}\label{Sec:DPMFramework}
DPMs approximate real data distributions $x_0\sim q(x)$ using two main processes: a \colorbox{cyan!25}{forward}\\ 
\colorbox{cyan!25}{diffusion process} and a \colorbox{magenta!25}{reverse diffusion process}. Our work mainly concerns the frameworks of \citet{sohl2015deep} and \citet{ho2020denoising}; and see more work based on the principle of diffusion in \citet{song2020denoising} and \citet{nichol2021improved}.

\subsection{The Framework}\label{Sec:TheFramework}

The \colorbox{cyan!25}{forward diffusion process} adds Gaussian noise to a sample from $q(x)$ in $T$ time-steps,
\begin{align}\label{Eq:DpmForward}
    q(x_t \lvert x_{t-1}) = \mathcal{N}(x_t; \sqrt{(1 - \beta_t)}x_{t-1}, \beta_t \mathbf{I})
\end{align}
where the magnitude of the noise is controlled by a pre-defined variance schedule $\{\beta_t\in(0, 1)\}_{t = 1}^T$. This results in a gradual loss of distinguishable features in the sample as $t$ increases. Furthermore, using properties of the Gaussian distribution, we can rewrite 
\begin{align}\label{Eq:xtx0}
    q(x_t \lvert x_0) = \mathcal{N}(x_t; \sqrt{\Bar{\alpha}_t}x_0, (1 - \Bar{\alpha}_t) \mathbf{I})
\end{align}
where $\alpha_t = 1 - \beta_t$ and $\Bar{\alpha}_t = \prod_{s = 1}^t \alpha_s$.

Then, a \colorbox{magenta!25}{reverse diffusion process} removes noises to synthesise a data as if it were sampled from the real data distribution $q(x)$. A model $p_\theta$ with weights $\theta$ is learned to approximate the conditional probabilities (via mean $\mu$ and covariance $\Sigma$) between $q(x)$ and the Gaussian noise input $x_T$
\begin{align}
    p_\theta(x_{0:T}) = p(x_T)\prod_{t = 1}^{T} p_\theta(x_{t - 1}\lvert x_t) \hspace{1mm}\text{ and }\hspace{1mm} p_\theta(x_{t - 1}\lvert x_t) = \mathcal{N}(x_{t - 1}; \mu_\theta(x_t, t), \Sigma_\theta(x_t, t)).
\end{align}

\subsection{Sampling and Training}
The forward process is tractable when conditioned on $x_0$:\\
\begin{align}
    &q(x_{t - 1}\lvert x_t, x_0) = \mathcal{N}(x_{t - 1}; \Tilde{\mu}_t(x_t, x_0), \Tilde{\beta}_t\mathbf{I}),\\
    &\text{where }\hspace{1mm} \Tilde{\mu}_t(x_t, x_0) = \frac{\sqrt{\Bar{\alpha}_{t - 1}}\beta_t}{1 - \Bar{\alpha}_t}x_0 + \frac{\sqrt{\alpha_t}(1 - \Bar{\alpha}_{t - 1})}{1 - \Bar{\alpha}_t}x_t \hspace{1mm}\text{ and }\hspace{1mm} \Tilde{\beta}_t = \frac{1 - \Bar{\alpha}_{t - 1}}{1 - \Bar{\alpha}_t}\beta_t,
\end{align}
and \citet{ho2020denoising} showed a DPM should learn to configure
\begin{align}\label{Eq:DpmLearningMu}
    \mu_\theta(x_t, t) = \frac{1}{\sqrt{\alpha_t}}\left(x_t - \frac{\beta_t}{\sqrt{1 - \Bar{\alpha}_t\epsilon_\theta(x_t, t)}}\right)
\end{align}
to predict the added noise $\epsilon$ in $x_t$ at time $t$ in Equation (\ref{Eq:DpmForward}) via approximating $\epsilon_\theta(x_t, t)$.

\begin{algorithm}[t]
\caption{Training the DPM with a U-Net backbone.}\label{Alg:Training}

\begin{algorithmic}[1]
    \STATEnonum \textbf{Initialises network parameters}: U-Net's $\theta$

    \State \textbf{repeat}
    \STATEnonum \hspace*{5mm}\underline{\# Prior to U-Net}
    
    \State \hspace*{5mm}
    $\xi_\text{0} \leftarrow \mathfrak{D}_\text{real}$
    \hspace*{64mm}Sample from real data $\mathfrak{D}_\text{real}$

    \State \hspace*{5mm}
     $t\sim \texttt{Uniform}(\{1,\ldots,T\})$
    \hspace*{40mm}Sample noise level

    \State \hspace*{5mm} 
    Reformulate $\xi_\text{0}$ to acquire $x_\text{0}$, 
    \hspace*{37.75mm}see Section \ref{Sec:DataFormulation}

    \State \hspace*{5mm}
    Inject noise $\epsilon$ in $x_\text{0}$ to acquire $x_t$, 
    \hspace*{32.75mm}see Equation (\ref{Eq:IntroducingManualNoise})

    \STATEnonum\textcolor{white}{.}
    \STATEnonum \hspace*{5mm}\underline{\# Applying U-Net}

    \State \hspace*{5mm}
    Estimate noise $\epsilon_\theta$ with the U-Net, 
    \hspace*{31.875mm}see Section \ref{Sec:Module}

    \State \hspace*{5mm}
    Estimate the one-step reconstruction $\hat{x}_0$ using  $\epsilon_\theta$, 
    \hspace*{10.25mm}see Equation (\ref{Eq:OneStepS2})

    \State \hspace*{5mm}
    Compute the noise estimation loss $\mathcal{L}_\text{Noise}$ with $\epsilon_\theta$, 
    \hspace*{9.875mm}see Equation (\ref{Eq:Loss})

    \State \hspace*{5mm}
    Compute the reconstruction losses $\mathcal{L_\text{Recon}}$ with $\hat{x}_0$, 
    \hspace*{8.75mm}see Section \ref{Sec:AuxLoss}

    \State \hspace*{5mm}
    Take a gradient step on $\nabla_\theta\mathcal{L}_\text{Tot} = \nabla_\theta(\mathcal{L_\text{Noise}} + \mathcal{L_\text{Recon}})$  

    \State \textbf{until converge}
    
\end{algorithmic}
\end{algorithm}

\begin{algorithm}[t]
\caption{Sampling for synthetic data $\mathfrak{D}_\text{Syn}$.}\label{Alg:Sampling}

\begin{algorithmic}[1]
    \STATEnonum 
    \STATEnonum \underline{\# Begin with random time series}
    
    \State
    $x_T \sim \mathcal{N}(\mathbf{0}, \mathbf{I})$

    \STATEnonum\textcolor{white}{.}
    \STATEnonum \underline{\# Iterative denoising with Langevin dynamics}

    \State
    \textbf{for} $t = T, \ldots, 1$ \textbf{do}

    \State \hspace*{5mm}
     $z \sim \mathcal{N}(\mathbf{0}, \mathbf{I})$

    \State \hspace*{5mm} 
    $x_{t - 1} = \frac{1}{\sqrt{\alpha_t}}\left(x_t - \frac{\beta_t}{\sqrt{1 - \Bar{\alpha}_t\epsilon_\theta(x_t, t)}}\right) + \sigma_t\mathbf{z}$,
    \hspace*{17.375mm} see Equation (\ref{Eq:ReverseProcess})

    \State
    \textbf{end for} 

    \State \textbf{return} $x_0$ as $\mathfrak{D}_\text{Syn}$
    
\end{algorithmic}
\end{algorithm}

The perturbed inputs from Equation (\ref{Eq:xtx0}) can be rewritten as 
\begin{align}\label{Eq:IntroducingManualNoise}
    x_t(x_0, \epsilon) = \sqrt{\Bar{\alpha_t}}x_0 + \sqrt{1 - \Bar{\alpha_t}}\epsilon \hspace{1mm}\text{ for }\hspace{1mm} \epsilon\sim \mathcal{N}(\mathbf{0}, \mathbf{I}).
\end{align}
Combined with the $\epsilon_\theta(x_t, t)$ in Equation (\ref{Eq:DpmLearningMu}), \citet{ho2020denoising} further showed that optimising
\begin{align}\label{Eq:Loss}
    \mathcal{L}_\text{Noise}(\theta) = \mathbb{E}_{t, x_0, \epsilon}\left[ \lVert \epsilon - \epsilon_\theta\left(\sqrt{\Bar{\alpha}_t}x_0 + 
                         \sqrt{1 - \Bar{\alpha}_t}\epsilon, t\right) \lVert^2 \right] 
\end{align}
is equivalent\footnote{
The equivalent loss is actually
$\mathbb{E}_{x_0, \epsilon}\left[ \frac{\beta_t^2}{2\sigma_t^2\alpha(1 - \Bar{\alpha}_t)} \lVert \epsilon - \epsilon_\theta\left(\sqrt{\Bar{\alpha}_t}x_0 + 
                         \sqrt{1 - \Bar{\alpha}_t}\epsilon, t\right) \lVert^2 \right]$
with the additional coefficients. However \citet{ho2020denoising} noted that it was beneficial to train without the additional coefficients.
} to optimising the negative log-likelihood using the variational lower bound.

To create novel data, we followed \citet{song2019generative}'s score-based generative method using Langevin dynamics with modifications on Equation (\ref{Eq:DpmLearningMu}), sampling $\mathbf{z}\sim\mathcal{N}(\mathbf{0}, \mathbf{I})$ where $\sigma^2_t = \beta_t$ and
\begin{align}\label{Eq:ReverseProcess}
    x_{t - 1} = \frac{1}{\sqrt{\alpha_t}}\left(x_t - \frac{\beta_t}{\sqrt{1 - \Bar{\alpha}_t\epsilon_\theta(x_t, t)}}\right) + \sigma_t\mathbf{z}. 
\end{align}

We provide supporting pseudo-codes that outline the structure of our proposed method in Algorithms \ref{Alg:Training} and \ref{Alg:Sampling}. Details within the pseudo-codes will be further discussed in subsequent sections.

\section{Ground Truth Datasets}\label{Sec:TruthData}
We based our work on the Health Gym project~\citep{kuo2022health}, which used GANs to generate synthetic longitudinal data from two health-related databases:
MIMIC-III~\citep{johnson2016mimic} and EuResist~\citep{zazzi2012predicting}. The authors used these databases to generate synthetic datasets for the management of acute hypotension and human immunodeficiency virus (ART for HIV). The patient cohorts were defined using inclusion and exclusion criteria from previous studies: \citet{gottesman2019guidelines} for acute hypotension and \citet{parbhoo2017combining} for ART for HIV.
The generated datasets include a comprehensive set of variables that can be utilised as observations, actions, and rewards in RL problems
aimed at managing patient illnesses. See all variables in Appendix \ref{App:Variables}.

\textbf{Acute Hypotension}\\
This dataset was extracted from MIMIC-III and was originally proposed by \citet{gottesman2019guidelines}. It comprises of 3,910 patients with 48-hour clinical variables, aggregated per hour in the time-series. The dataset includes variables with suffix (M) to indicate the measurement at a specific point in time and is significant due to its informativeness in missing data in clinical time series, which can indicate the need for laboratory tests. In their work, \citeauthor{gottesman2019guidelines} utilised this dataset to develop an RL agent that suggested optimal fluid boluses and vasopressors for acute hypotension management, with actions being made in a discrete action space by binning the boluses and vasopressors into multiple categories. Refer to Table \ref{Tab:Hypotension} for more details.

\textbf{ART for HIV}\\
The real HIV dataset is based on a cohort of individuals from the EuResist database, as proposed by \citet{parbhoo2017combining}. The study employs a mixture-of-experts approach for therapy selection, utilising kernel-based methods to identify clusters of similar individuals and an RL agent to optimise treatment strategy. The dataset consists of 8,916 individuals who started therapy after 2015 and were treated with the 50 most common medication combinations, including 21 different types of medications. Demographics, viral load (VL), CD4 counts, and regimen information are included in the dataset. The length of therapy in the dataset varies, thus the records were truncated and modified to the closest multiples of 10-month periods, resulting in a shortest record length of 10 months and a longest record length of 100 months, each summarising patient observations over a 1-month time period. Again, we include binary variables with the suffix (M) to indicate whether a variable was measured at a specific time. Refer to Table \ref{Tab:HIV} for more details.

\section{Methods}\label{Sec:Methods}
This section details the setups for generating mixed-type time-series data with DPMs.

\subsection{Data Formulation for Mixed-Type Inputs \& Outputs}\label{Sec:DataFormulation}
For each iteration, we draw ground truth data \colorbox{blue!25}{$\xi_0$} from the set of clinical datasets (see Section \ref{Sec:TruthData}), and reformulate it to \colorbox{purple!25}{$x_0$} (to be addressed below). We also select a noise level $t$ and its corresponding strength of perturbation $\beta_t$ to introduce corruption to \colorbox{purple!25}{$x_0$} following Equation (\ref{Eq:xtx0}) to acquire the noisy inputs $x_t$. To estimate the manually injected noise $\epsilon$ of Equation (\ref{Eq:IntroducingManualNoise}), we feed $x_t$ into a tailored implementation of U-Net~\citep{ronneberger2015u}, which serves as our backbone network for the denoising operations. The output of the U-Net network $\epsilon_\theta$ is the predicted estimation for $\epsilon$. 

Our datasets encompass numeric, binary, and categorical variables. Hence, we elaborate on the data formulation prior to presenting it to the model. The ground truth data is partitioned as\\
$\colorbox{blue!25}{$\xi_0$}= \colorbox{blue!10}{$\xi_{0, \text{[num]}}$} \oplus \colorbox{blue!10}{$\xi_{0, \text{[alt]}}$}$, with the numeric subset \colorbox{blue!10}{$\xi_{0, \text{[num]}}$} and the non-numeric subset \colorbox{blue!10}{$\xi_{0, \text{[alt]}}$}. We transform each numeric feature in \colorbox{blue!10}{$\xi_{0, \text{[num]}}$} to the range $\in [0, 1]$ and derive \colorbox{purple!10}{$x_{0, \text{[num]}}$}. Each non-numeric variable \colorbox{blue!10}{$\xi_{0, \text{[alt]}}^{(i)}$} is converted into a list of one-hot vectors where the observed class is assigned a value of 1 and the others a value of 0. Here are some examples:
\begin{itemize}[noitemsep, topsep=0pt]
    \item For the binary variable \texttt{Gender = Female}, we have\\
    $\texttt{Gender} =
        \begin{bmatrix}
            \texttt{Female}\\
            \texttt{Male}
        \end{bmatrix} =
        \begin{bmatrix}
            1\\
            0
        \end{bmatrix}
        $, and
    \item for the categorical variable \texttt{Ethnicity = African}, we have\\
    $\texttt{Ethnicity} =
        \begin{bmatrix}
            \texttt{Asian}\\
            \texttt{African}\\
            \texttt{Caucasian}\\
            \texttt{Other}
        \end{bmatrix} =
        \begin{bmatrix}
            0\\
            1\\
            0\\
            0
        \end{bmatrix}
        $.
        \end{itemize}
We denote the aggregate of all one-hot vectors as $\colorbox{purple!10}{$x_{0, \text{[alt]}}$} = \bigcup_i \text{OneHot}\left(\colorbox{blue!10}{$\xi_{0, \text{[alt]}}^{(i)}$}\right)$; and that\\
$\colorbox{purple!25}{$x_0$}= \colorbox{purple!10}{$x_{0, \text{[num]}}$} \oplus \colorbox{purple!10}{$x_{0, \text{[alt]}}$}$. Embeddings~\citep{landauer1998introduction, mikolov2013efficient, mottini2018airline} are not necessary in our framework. The forward diffusion process of the DPM (refer to Equation~(\ref{Eq:IntroducingManualNoise})) directly applies noise to the one-hot vectors.

At test time, we randomly sample a noisy input $x_T$ from a Gaussian distribution. Then, we iteratively estimate the corruption $\epsilon_\theta(x_t, t)$ at step $t$ using our U-Net backbone to generate a less noisy $x_{t - 1}$ as per Equation~(\ref{Eq:ReverseProcess}). Once\footnote{Similar to \citet{ho2020denoising}, we used $x_0$ and $\xi_0$ for both the clean inputs and the denoised, reconstructed output; the terms should thus be used in context of before and after the reverse diffusion process.} we reach the allegedly clean and novel data \colorbox{purple!25}{$x_0$}, we compartmentalise it into \colorbox{purple!10}{$x_{0, \text{[num]}}$} and \colorbox{purple!10}{$x_{0, \text{[alt]}}$} to reverse the transformation in \colorbox{purple!10}{$x_{0, \text{[num]}}$}, resulting in \colorbox{blue!10}{$\xi_{0, \text{[num]}}$}. Next, we employ softmax to recover the non-numeric variables such that $\colorbox{blue!10}{$\xi_{0, \text{[alt]}}$} = \bigcup_i \text{Softmax}\left(\colorbox{purple!10}{$x_{0, \text{[alt]}}^{(i)}$}\right)$.

The dimensionality of the noisy input is $x_t \in \mathbb{R}^{\mathfrak{B} \times 1 \times \mathfrak{L} \times \mathfrak{N}}$, where $\mathfrak{B}$ corresponds to the batch size,  $\mathfrak{L}$ denotes the length of the time-series, and that there is 1 feature channel for all the $\mathfrak{N}$ variables. In Section \ref{Sec:TruthData}, it was mentioned that all acute hypotension data possess a fixed sequence with a length of 48 units, hence $\mathfrak{L}_\text{hypotension} = 48$. On the other hand, the HIV data has variable lengths and we utilise zero-padding to bring all the data to a pre-defined maximal length of $\mathfrak{L}_\text{HIV} = 100$. This setup hence obviates the need for curriculum learning~\citep{bengio2009curriculum} to enable training.

Moreover, inferring the size $\mathfrak{N}$ is a straightforward task, as it solely involves concatenating the numeric and one-hot representations of the binary and categorical variables in $x_t$. To illustrate, consider the ART for HIV dataset, whose variable specifications are provided in Table~\ref{Tab:HIV}. By summing up the corresponding levels of every variable (with 1 for numeric variables), we obtain that\\
\hspace*{5mm}$\mathfrak{N}_\text{HIV} = \text{sum}(\{1, 1, 1, 2, 4, 6, 4, 4, 6, 2, 2, 2, 2\}) = 37$.\\
Likewise, we deduce that $\mathfrak{N}_\text{hypotension} = 54$ as per its respective specifications in Table \ref{Tab:Hypotension}.

\begin{figure}[t]
    \centering
    \includegraphics[width = 0.85\linewidth]{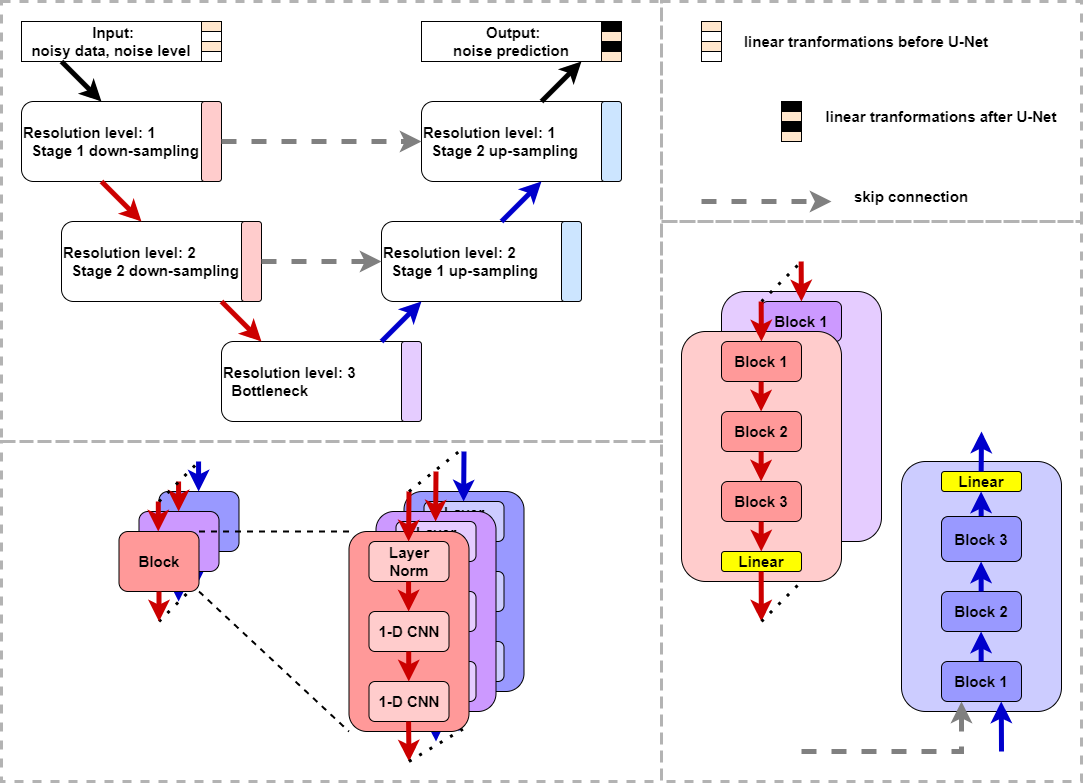}
    \caption{\label{Fig:UNet}An overview of the elements of our U-Net.\\
    Our U-Net is depicted in the top left panel, with the down-sampling, bottleneck, and up-sampling procedures denoted by the colors red, purple, and blue, respectively. Top right: The presence of linear transformations for pre- and post-processing (see Section \ref{Sec:Module}-d)). Bottom right: The local features in each resolution level is processed with block processing units and linear transformations (see Section \ref{Sec:Module}-d)). Bottom left: the anatomy shared by all blocks (see Section \ref{Sec:Module}-c)).}
\end{figure}

\subsection{The U-Net Backbone}
U-Net~\citep{ronneberger2015u} is a convolutional neural network (CNN) architecture originally developed for medical image segmentation. As shown in Figure \ref{Fig:UNet}, the architecture has many details. The down-sampling~\citep{zeiler2014visualizing} compartment extracts high-level features from noisy data, while skip connections~\citep{venables1999modern} maintain fine-grained details and spatial information. The up-sampling compartment estimates noise for reconstructing clean data, leveraging localised features via the skip connections. U-Net is especially useful in denoising spatially correlated noise of varying intensities; and has been employed in various DPM applications~\citep{saharia2022palette, ho2022video, li2022srdiff}.

\subsubsection{The Modules}\label{Sec:Module}
We depicted the U-Net processing procedure in Figure \ref{Fig:UNet}. Refer to hyper-parameters in Section \ref{Sec:Setup} and examples of the dimensionality change in intermediate neural activations in Section \ref{Sec:DimChange}.

\ref*{Sec:Module}-a): \underline{Embedding the noise level}\\
In Equation (\ref{Eq:DpmLearningMu}), the noise prediction process of DPM is enabled via $\mu_\theta$ to create $\epsilon_\theta$ to predict noise $\epsilon$. Notably, $\mu_\theta$ is informed by noise level $t$, which is used to iteratively estimate noise across various levels. To this end, we adopt the Transformer sinusoidal position embedding method~\citep{vaswani2017attention}, as applied in \citet{ho2020denoising}, to featurise the noise level. These noise level embeddings are then incorporated into the U-Net architecture, and are fed as input to each intermediate neural activation stage that arises from the down- and up-sampling operations.

\ref*{Sec:Module}-b): \underline{Down- and up-sampling}\\
All CNNs employed in our design are one-dimensional (1-D) and do not possess a causal architecture. Thus when we denoise the noisy acute hypotension datum $x_t \in \mathbb{R}^{\mathfrak{B} \times 1 \times \mathfrak{L} \times \mathfrak{N}}$ (see Section \ref{Sec:DataFormulation}) with a fixed length of $\mathfrak{L} = 48$, the U-Net could simultaneously denoise the noisy data at positions 10 and 20. Our U-Net hence processes data similar to the autoencoding style of BERT~\citep{devlin2019bert}, as opposed to the autoregressive style of GPT~\citep{radford2018improving} (\textit{i.e.,} we are not limited to denoising from left-to-right in a single~direction). See more discussion in Section \ref*{Sec:Module}-d).

\ref*{Sec:Module}-c): \underline{Block feature extractor}\\
After each stage of sampling operation, the noisy data is further processed while maintaining the same resolution level. Within each level, we utilise three successive feature extraction blocks, each composed of layer normalisation~\citep{ba2016layer} followed by two 1-D CNNs.

\ref*{Sec:Module}-d): \underline{Distinctive Additions to Our U-Net Architecture}\\
We found that the application of the 1-D CNNs alone is insufficient for denoising. As elaborated upon in Section \ref*{Sec:Module}-b), 1-D CNNs have the capability to denoise the noisy data simultaneously at positions 10 and 20, but for each feature independently. For ART for HIV, denoising VL is hence done independently of the regimen taken. While 2-D CNNs may seem more viable, an incorrect kernel size can still cause the erroneously denoising the of \{VL, regimen\} (in the kernel), while leaving out the relevant information of \{CD4, Ethnicity\} (out of the kernel). The need to concurrently denoise multiple time-series variables introduces a level of complexity that is not encountered in the DPM's application in speech~\citep{lu2022conditional}. 

This can be addressed by applying additional linear transformation layers on the $\mathfrak{N}$ dimension of $x_t$. As a consequence, the U-Net no longer denoises data at a variable level and instead denoises data on their latent features. Inspired by \citet{lin2013network}, we also include linear transformations to each up- and down-sampling 1-D CNN (see the bottom right panel of Figure \ref{Fig:UNet}) to process local patches within the receptive field.

Additional linear transformations are then employed on the final up-sampling output. This restructures the predicted noise made on the latent structure back to the $\mathfrak{N}$ sequences on $x_t$.

\subsection{Auxiliary Loss Functions}\label{Sec:AuxLoss}

Capturing both individual variable fidelity and correlations is crucial for realistic synthetic EHR. Previous studies in GANs used auxiliary loss functions, such as separate encoders and matching loss~\citep{li2021generating}, or assessed real data correlation before training~\citep{kuo2022health}. However, these approaches synthesise data per iteration, making them computationally costly. Including such losses in DPM is prohibitively expensive due to DPM's slow sampling nature~\citep{xiao2021tackling}.

To circumvent the iterative denoising in Equation (\ref{Eq:ReverseProcess}), we estimate a one-step reconstruction loss
\begin{align}
    &\mathcal{L}_{\text{Recon}_1} = \lVert x_0 - \hat{x}_0(t, \epsilon_\theta) \lVert_2^2, \hspace{1mm}\text{ where}\label{Eq:OneStep}\\
    &\hat{x}_0(t, \epsilon_\theta) = \frac{x_t - \sqrt{(1 - \Bar{\alpha}_t)}\epsilon_\theta}{\sqrt{\Bar{\alpha}_t}}.\label{Eq:OneStepS2}
\end{align}
We utilise the predicted noise $\epsilon_\theta$ from the U-Net to replace the actual noise $\epsilon$, and apply a one-step reverse process to approximate the clean data $x_0$ using $\hat{x}_0$. This is analogous to taking a large step on the vector field~\citep{song2019generative, song2020score}, and may result in overshooting during the reconstruction dynamics. Hence, the use of $\hat{x}_0$ is restricted solely to our auxiliary loss.

We introduce a second auxiliary loss function to our model, defined as 
\begin{align}
    &\mathcal{L}_{\text{Recon}_2} = \lVert \mathfrak{U}(x_0) - \mathfrak{U}\left(\hat{x}_0(t, \epsilon_\theta)\right) \lVert_2^2, \hspace{1mm}\text{ where}\label{Eq:OneStepS3}\\
    &\mathfrak{U}(v) = \text{max}(0, v \hspace*{1mm}\mathcal{U}_1) \hspace*{1mm}\mathcal{U}_2.\label{Eq:OneStepS4}
\end{align}
At each iteration, we construct two random matrices $\mathcal{U}_1$ and $\mathcal{U}_2$ that do not require training. These matrices are used to project the clean data $x_0$ and its reconstructed counterpart $\hat{x}_0$ to a latent space, and the difference between them is minimised. This approach is inspired by \citet{salimans2016improved}'s mini-batch discrimination technique and aims to minimise the variability of the variable combinations in a randomly projected feature space.

\section{Experimental Setup}\label{Sec:Setup}
This section details the hyper-parameters implemented in our experiments and the performance metrics utilised to evaluate the quality of our generated dataset. Our evaluation involved comparing our synthetic dataset with the datasets presented in \citet{kuo2022health}. The lack of baseline models can be attributed to the limited amount of literature addressing the generation of longitudinal clinical datasets that possess mixed-type variables. Additionally, only the synthetic datasets provided by \citet{kuo2022health} were made publicly available for conducting comparisons.

Henceforth, in this manuscript, we employ the following notation:\\
\hspace*{5mm} $\mathfrak{D}_\text{real}$ to denote the ground truth dataset;\\
\hspace*{5mm} $\mathfrak{D}_\text{null}$ to refer to the synthetic dataset generated via \citet{kuo2022health}'s GAN; and\\
\hspace*{5mm} $\mathfrak{D}_\text{alt}$ to represent the alternative synthetic dataset generated using our DPM setup.

\subsection{Hyper-parameters}\label{Sec:DimChange}
\textbf{Hyper-parameters for the U-Net}\\
Following Section \ref{Sec:Module}-d), we choose to linearly project the $\mathfrak{N}$ variables in the input to a latent space of dimensionality 256. After this preliminary step, we employ our U-Net for denoising.

As detailed in Section \ref{Sec:Module}-b), we adopt 3 distinct resolution levels. More specifically, resolution level 1 maintains the initial length of the noisy sequence for all time-series data, whereas the succeeding resolution levels condense the sequences while augmenting feature dimensions. In resolution level 2, all time-series have feature size 10, and in resolution level 3, all time-series possess feature size 20, regardless of the underlying dataset. However, the alteration in the length of the time-series depended on the dataset. For acute hypotension, resolution level 1 sequences span 48 time steps, which are subsequently reduced to 12 and 3 in resolution levels 2 and 3, respectively. Likewise in ART for HIV, they change from length 100 to 10 and then 3.

Following the previous descriptions, the 1-D CNNs employed in the blocks of Section \ref{Sec:Module}-c) possess feature dimensions of 10 and 20 at resolution levels 1 and 2 respectively. Whereas the features in the bottleneck of resolution level 3 reduces from 20 to 10, subsequently reverting to 20.

\textbf{An example using acute hypotension}\\
The input data $x_t\in\mathbb{R}^{\mathfrak{B} \times 1 \times 48 \times 37}$ comprises a sequence length $\mathfrak{L}$ of 48 and $\mathfrak{N}$ variables of 37 (see Section \ref{Sec:DataFormulation}). We project and contruct the latent structure of the noisy data in $\mathbb{R}^{\mathfrak{B} \times 1 \times 48 \times 256}$. In the subsequent use of U-Net, the dimensionality transforms to $\mathbb{R}^{\mathfrak{B} \times 10 \times 12 \times 256}$ in resolution level 2 and then $\mathbb{R}^{\mathfrak{B} \times 20 \times 3 \times 256}$ in resolution level 3.

\textbf{Hyper-parameters for the DPM \& Optimisation}\\
We set the maximum perturbation at $\beta_0 = 0.01$ and the minimum at $\beta_T = 1\times10^{-4}$ (see Equation (\ref{Eq:DpmForward})) across both datasets. However, we use $T = 1000$ for acute hypotension and $T = 500$ for ART for HIV. The intermediate perturbations $\beta_t$ are distributed uniformly across the $T$ levels. As previously stated in Section \ref{Sec:Module}-a), the denoising procedure of our DPM is informed by the noise level $t$. This information is conveyed to the U-Net architecture as a Transformer sinusoidal position embedding, featuring an embedding dimensionality of $100$.

Our DPMs are updated using the Adam optimiser~\citep{kingma2014adam} with learning rate $1\times10^{-3}$. We employ a batch size of $128$ for the acute hypotension and ART for HIV. The DPMs are trained for $5000$ epochs for acute hypotension and $3000$ epochs for ART for HIV. In addition, the losses are weighted at a ratio of $1:20:10$ for $\mathcal{L}_\text{noise}$, $\mathcal{L}_{\text{Recon}_1}$, and $\mathcal{L}_{\text{Recon}_2}$ (see Section \ref{Sec:AuxLoss}), respectively.

\subsection{Metrics}\label{Sec:Metric}

We put forth five desiderata:
\begin{itemize}[noitemsep, topsep=0pt]
    \item Section \ref{Sec:IndReal}: that all generated variables to exhibit individual realism;
    \item Section \ref{Sec:CorrAna}: that the collective realism of all variables hold across time;
    \item Section \ref{Sec:Diversity}: that there exists a sufficiently high level of diversity in variables;
    \item Section \ref{Sec:SecEst}: that our synthetic datasets ensure patient privacy; and
    \item Section \ref{Sec:UtilityInves}: that our datasets can function as a substitute for a genuine dataset in\\
    \hspace*{21mm}downstream model construction.
\end{itemize}

\subsubsection{Assessing Individual Realisticness}\label{Sec:IndReal}
We leverage two plots to assess the individual realisticness. For numeric variables, we use kernel density estimations (KDEs) \citep{davis2011remarks} to overlay the synthetic distribution on top its genuine counterpart. For binary and categorical variables, we use side-by-side barplots to demonstrate the percentage share of each level.

\begin{wrapfigure}{r}{0.4\textwidth}
    \centering
    \vspace*{-2mm}
    \includegraphics[width=\linewidth]{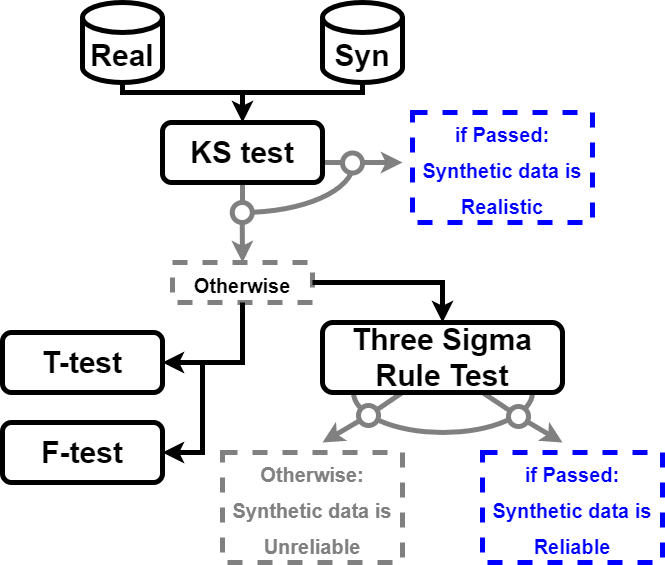}
    \caption{\label{Fig:HST}Statistical tests.\\
    The sequence of the hypothesis tests.}
    \vspace*{-5mm}
\end{wrapfigure}

Following  \citet{kuo2022health} and \citet{hernadez2023synthetic}, we perform four statistical tests on the synthetic datasets shown in Figure \ref{Fig:HST}. We begin with the two-sample Kolmogorov-Smirnov (KS) test~\citep{hodges1958significance} to evaluate whether the synthetic variables effectively capture the distributional characteristics of their real counterparts. If a synthetic variable passes the KS test, it is deemed to be realistic and can be considered as having been drawn from the real datasets. Otherwise, we seek to identify the underlying reasons for its lack of realism.

The perceived lack of realism could be understood using the Student's t-test~\citep{yuen1974two} and the F-test. Snedecor's F-test~\citep{snedecor1989statistical} is used for numeric variables; and we use the analysis of variance F-test for binary and categorical variables. The t-test verifies the alignment between means, while the F-test assesses the agreement in variances. However, in the event that a synthetic variable fails the KS test, neither the t-test nor the F-test can be used to assess the reliability of the synthetic variable. Hence, we choose the three sigma rule test~\citep{pukelsheim1994three} (by default, with 2 standard deviations) to evaluate whether the synthetic values fall within a plausible range of real variable values.

In contrast to image generation, we cannot employ the inception score (IS)~\citep{salimans2016improved} and the Fr\'echet inception distance (FID)~\citep{heusel2017gans} to evaluate the quality of our generated data. These metrics rely on the Inception v3 model~\citep{szegedy2015going}, which is unsuitable for analysing our longitudinal EHR data. Therefore, we follow the lead of \citet{goncalves2020generation} and add the Kullback-Leibler (KL) divergence as a complementary measure to estimate the similarity between the synthetic and real data distributions for a given variable.

We start with a preparation stage in which we bin each numeric variable into 20 equivalent classes. Then, for each discretised numeric variable, binary variable, and categorical variable, we calculate the KL divergence between the true distribution $q(x)$ and the learned distributions $p_\theta(x)$
\begin{align}\label{Eq:KldDiscrete}
\text{KLD}(p_\theta(x) \lVert q(x)) = \sum_{x\in X}p_\theta(x)\text{log}\frac{p_\theta(x)}{q(x)}.
\end{align}
Since KL divergence is defined at the variable level, we apply it to each variable individually on the synthetic datasets $\mathfrak{D}_\text{null}$ and $\mathfrak{D}_\text{alt}$ and then determine how many variables have a lower (and hence better) score. While \citet{goncalves2020generation} aggregated all their individual KL divergences, their synthetic dataset only contained categorical variables. Therefore, we find it beneficial to compare the KL divergence on a case-by-case basis for our mixed-type variables.

\subsubsection{Correlation Analysis}\label{Sec:CorrAna}

We employ Kendall's $\tau$ rank correlation~\citep{kendall1945treatment} to assess the relationships among variables in the mixed-type datasets. The correlation is computed in two ways: first, we calculate the \textit{static} correlations among all data points under the classic setup; second, we estimate the average \textit{dynamic} correlations following the approach of \citet{kuo2022health}.

The dynamic correlation is computed in two stages. Initially, we decompose each variable $i$ of every patient $\mathfrak{p}$ into a trend and a cycle using linear deconstruction, as shown below:
\begin{align}
X^{(i)}_{\mathfrak{p}} &= \text{Trend}\left(X^{(i)}_\mathfrak{p}\right) + \text{Cycle}\left(X^{(i)}_\mathfrak{p}\right).
\end{align}
Trends reveal macroscopic patterns in the time series data, such as overall increasing or decreasing trends, while cycles help us understand information on the microscopic level, such as periodic behaviours. After detrending the variables, we calculate the correlations separately for the trends and cycles and then average the values across all patients.

\subsubsection{Evaluating Diversity on the Data Structure}\label{Sec:Diversity}
To assess the level of diversity present in our synthetic datasets, we employ two metrics: the log-cluster metric $U$~\citep{woo2009global} and category coverage (CAT) proposed in \citet{goncalves2020generation}. The former, formulated as
\begin{align}
U = \text{log}\left(\frac{1}{\Gamma}\sum_{k = 1}^\Gamma\left[\frac{n_{k_\text{real}}}{n_{k}} - \frac{n_{k_\text{real}}}{n_{k_\text{real}} + n_{k_\text{syn}}}\right]^2\right),
\end{align}
measures the difference in latent structures between the real and synthetic datasets. To compute $U$, we first sample records from both the real and synthetic datasets and then merge the sub-datasets to perform a cluster analysis via k-means with $\Gamma = 20$ clusters. Here, $n_k$ represents the total number of records in cluster $k$, while $n_{k_\text{real}}$ and $n_{k_\text{syn}}$ denote the number of real and synthetic records in cluster $k$, respectively. We repeat this process 20 times for each synthetic dataset, with each repetition involving a sample of 100,000 real and synthetic records. A lower $U$ score indicates that the synthetic datasets are more realistic.

The latter metric, CAT, is defined as
\begin{align}
\text{CAT} = \frac{1}{\mathscr{U}}\sum_{u = 1}^{\mathscr{U}}\frac{\lVert \mathfrak{D}^{(u)}_{\text{syn}} \lVert}{\lVert \mathfrak{D}^{(u)}_{\text{real}} \lVert},
\end{align}
where $\mathscr{U}$ is the total number of binary and categorical variables, and $\mathfrak{D}^{(u)}_{\text{real}}$ and $\mathfrak{D}^{(u)}_{\text{syn}}$ represent the real and synthetic datasets, respectively, for the $u$-th variable. Specifically, CAT measures the completeness of the non-numeric classes in the synthetic datasets; it is the higher the better.

\subsubsection{Security Estimation}\label{Sec:SecEst}
We conduct two tests. First, we examine the minimum Euclidean distance between synthetic and actual records and verify that it is greater than zero, thus preventing any real records from being leaked into the synthetic dataset. Then, we utilise the sample-to-population attack in \citet{el2020evaluating} to assess the potential risk of an attacker learning new information by linking an individual in the synthetic dataset to the actual dataset.

The sample-to-population attack involves \textit{quasi-identifiers}, which are variables that may reveal an individual's identity, such as \texttt{Gender} and \texttt{Ethnicity} for the ART for HIV dataset. \textit{Equivalent classes} are then formed by combining these variables, resulting in groups such as \texttt{Male + Asian} and \texttt{Female + African}. The risk associated with linking a synthetic patient $s$ is estimated with
\begin{align}
\frac{1}{S}\sum_{s = 1}^{S}\left(\frac{1}{F_{s}} \times I_{s}\right),
\label{Eq:FakeRisk}
\end{align}
where $S$ represents the total number of records in the synthetic dataset, $I_{s} \in {0, 1}$ equals one if the equivalent class of synthetic $s$ is present in both datasets, and $F_{s}$ denotes the cardinality of the equivalent class in the actual dataset.

The \citet{european2014european} and \citet{canadian2019canadian} standards recommend that this risk should not exceed $9\%$ to balance synthetic data utility and security. By following these measures, we ensure that our synthetic datasets are secure and suitable for public use.

\subsubsection{Utility Investigation}\label{Sec:UtilityInves}
We employ both the synthetic and real datasets to train RL agents, and we consider a synthetic dataset to achieve a high level of utility if an RL agent trained on both real and synthetic datasets generates similar actions when presented with clinical conditions of patients.

We partitioned each dataset into a set of observational variables and a set of action variables. The observational variables describe the clinical condition of a patient, while the action variables define the actions an RL agent could take. We adopt the approach in \citet{liu2021offline} to reduce the observational dimensionality to five variables using cross decomposition~\citep{Wegelin00asurvey}. Next, we applied K-Means clustering~\citep{vassilvitskii2006k} with 100 clusters to define the state space and assigned each data point to their corresponding cluster label. The action space was defined as the set of unique values of the action variables.

Subsequently, we employed published reward functions to determine the optimal actions that an RL agent should take given a patient state\footnote{Refer to \citet{gottesman2019guidelines} and \citet{parbhoo2017combining} for the reward functions for acute hypotension and ART for HIV. In addition, see Sections 7.1 in the Appendix of \citet{kuo2022health} for additional details on the implementation for acute hypotension; and likewise Section 4.3.5 in \citet{kuo2022generating} for ART for HIV.}. We select batch-constrained Q-learning~\citep{fujimoto2019off} for utility investigation, and we update the policies for 100 iterations with a step size of 0.01.

\section{Experimental Results}\label{Sec:Results}
This section presents the results of the five desiderata outlined in Section \ref{Sec:Metric}.

\subsection{On the Individual Realisticness of the Variables}\label{Sec:IndFiel}
\begin{figure}[ht!]
    \centering
    \begin{subfigure}{\linewidth}
      \centering
      \includegraphics[width=\linewidth]{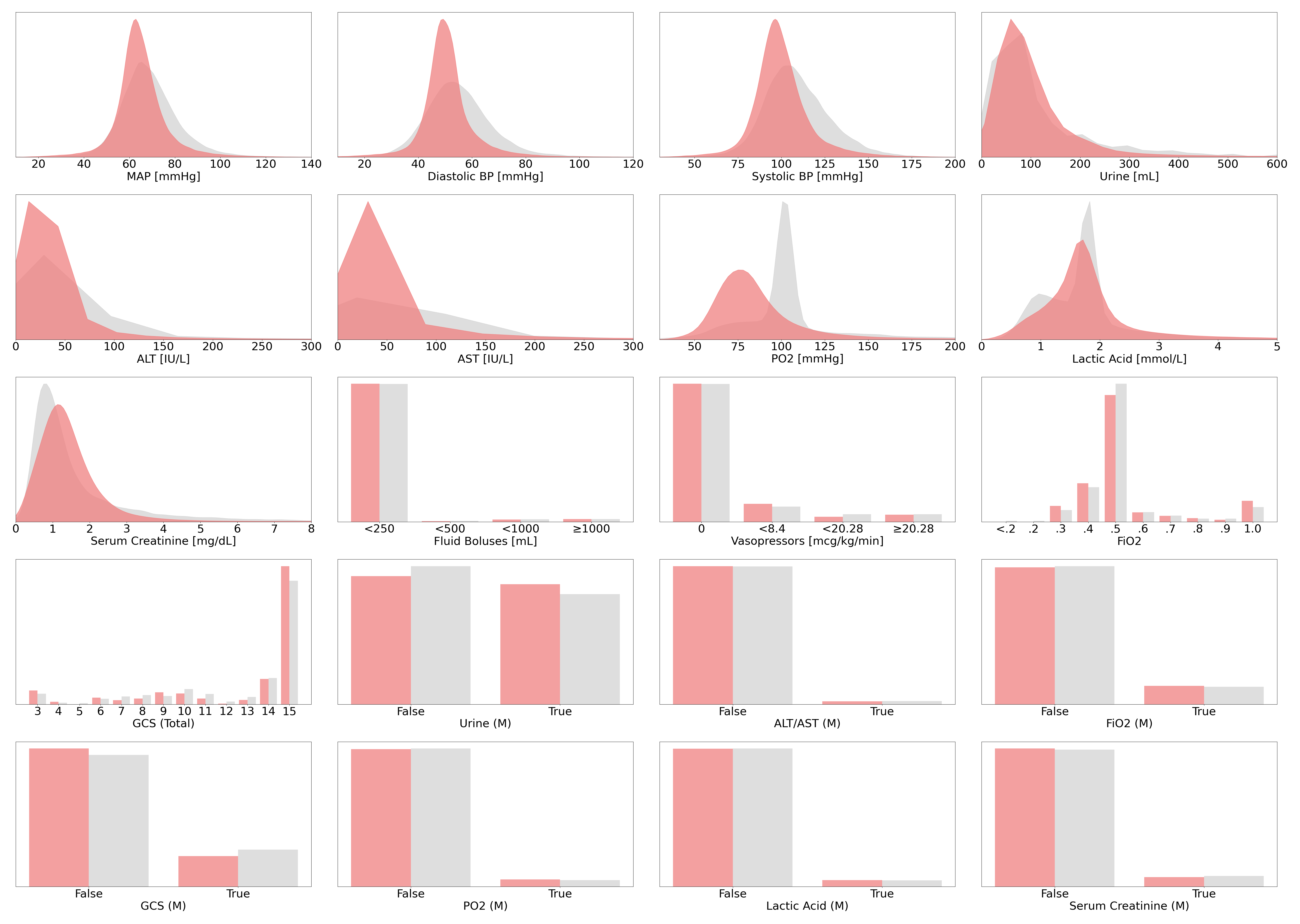}
      \caption{Synthetic dataset $\mathfrak{D}_\text{null}$ from \citet{kuo2022health} in pink.}
    \end{subfigure}
    
    \begin{subfigure}{\linewidth}
      \centering
      \includegraphics[width=\linewidth]{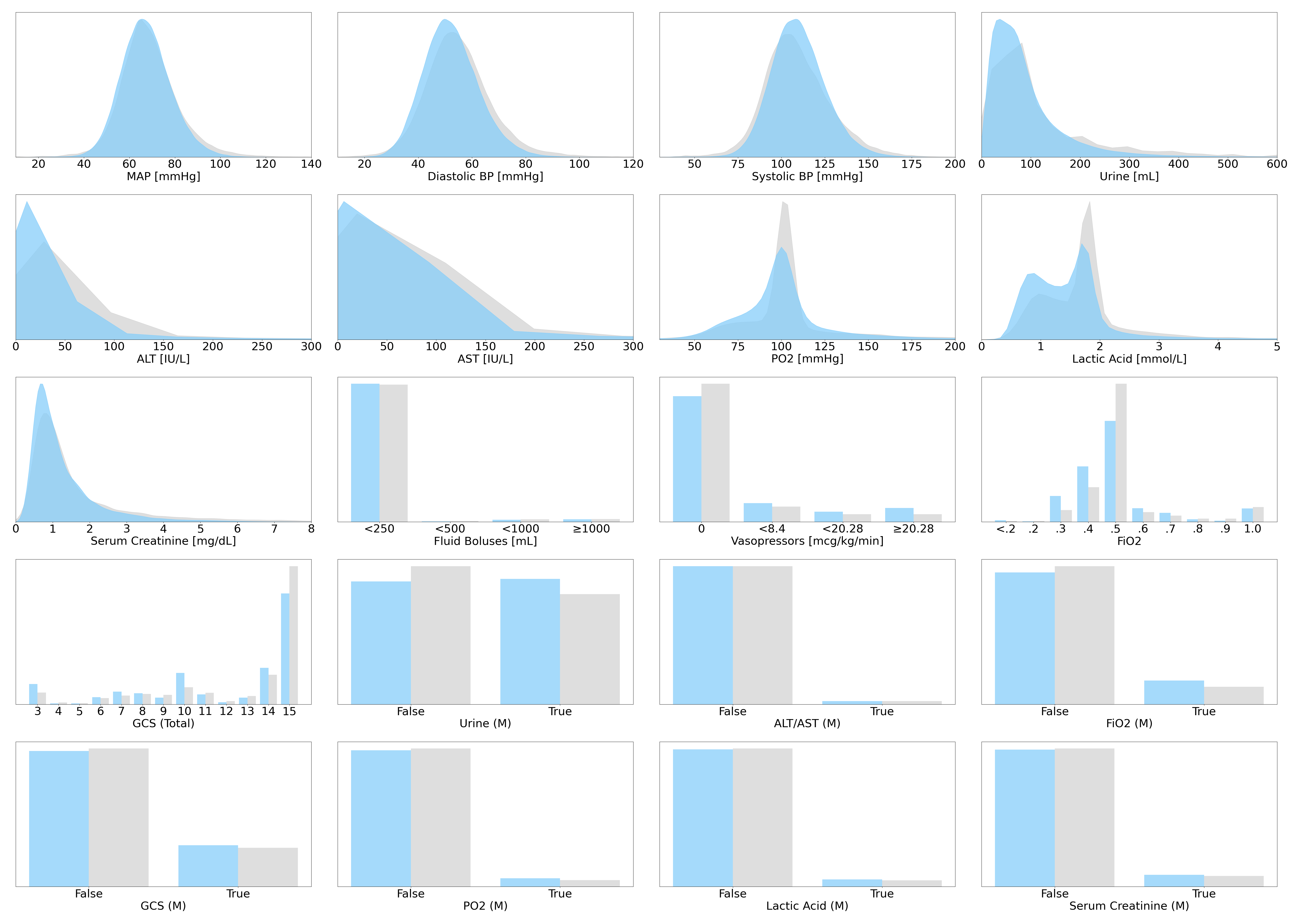}
      \caption{Synthetic dataset $\mathfrak{D}_\text{alt}$ from our DPM in blue.}
    \end{subfigure}%
    
    \caption{\label{Fig:KdeBarHpotension}Comparing the variables in acute hypotension, with those of $\mathfrak{D}_\text{real}$ in colour grey.}
\end{figure}

\textbf{Actute Hypotension}\\
The KDE plots\footnote{The kernel density estimation uses Gaussian kernels to estimate the probability density function of a continuous variable. Thus, the KDE function can potentially produce tails beyond the range of the data.
} 
and barplots for the individual variable comparisons are presented in Figure \ref{Fig:KdeBarHpotension}. The grey bars represent the real variables from $\mathfrak{D}_\text{real}$, while the respective pink and blue bars in subplots \ref{Fig:KdeBarHpotension}(a) and \ref{Fig:KdeBarHpotension}(b) depict the synthetic variables in $\mathfrak{D}_\text{null}$ and $\mathfrak{D}_\text{alt}$, generated using \citet{kuo2022health}'s Health Gym GAN and our DPM. Overall, the distributions in both subplots are comparable to their real counterparts in $\mathfrak{D}_\text{real}$. We observed that DPM captured the multi-modal nature of clinical variables better than GAN (\textit{e.g.,} \texttt{PaO$_2$} and \texttt{Lactic Acid}), but we also found that our DPM generated more instances of less common classes in \texttt{FiO$_2$}.

The synthetic variables in our DPM-generated hypotension dataset $\mathfrak{D}_\text{alt}$ are representative of their real counterparts in $\mathfrak{D}_\text{real}$. The statistics in Table \ref{Tab:HypotensionStats} in Appendix \ref{App:Stats} revealed that all variables passed the three sigma rule test and are reliable. Most variables passed the KS test and thus captured detailed information in the real distributions. The minority of variables that failed the KS test still passed the t-test and F-test, demonstrating that both the mean and the variance are captured and only missing the extreme details in the cumulative distribution function.

The KL divergences in Table \ref{Tab:HypotensionKLD} in Appendix \ref{App:Stats} indicated that most variables simulated by the DPM are on-par with those generated using GAN. Only the KL divergence of \texttt{FiO$_2$} was much larger in $\mathfrak{D}_\text{alt}$, consistent with the previous finding that our DPM simulated less common classes for \texttt{FiO$_2$}.

\textbf{ART for HIV}\\
Refer to all results in Appendix \ref{App:Stats}. In Figure \ref{Fig:KdeBarHIV}, we observed that the variable distributions in our DPM-generated $\mathfrak{D}_\text{alt}$ capture the features of the real variables more accurately than those in $\mathfrak{D}_\text{null}$ generated using GAN in \citet{kuo2022health}. The statistical tests reported in Table \ref{Tab:HIVStats} indicated that while all DPM-simulated variables are reliable, only \texttt{VL} failed the KS test. However, \texttt{VL} still passed the three sigma rule test; and that the KL divergences in Table \ref{Tab:HIVKLD} revealed that the quality of DPM-simulated distributions are on-par or superior to those generated using GAN.

\newpage

\newpage
\begin{figure}[ht!]
    \centering
    \begin{subfigure}{\linewidth}
      \centering
      \includegraphics[width=\linewidth]{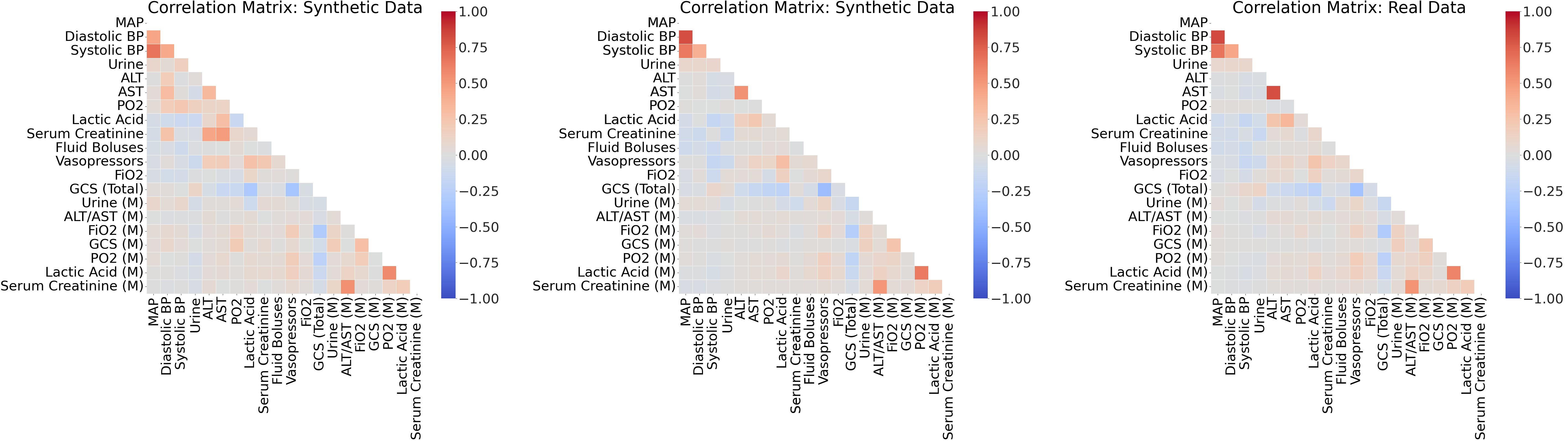}
      \caption{The static correlations.}
    \end{subfigure}
    
    \begin{subfigure}{\linewidth}
      \centering
      \includegraphics[width=\linewidth]{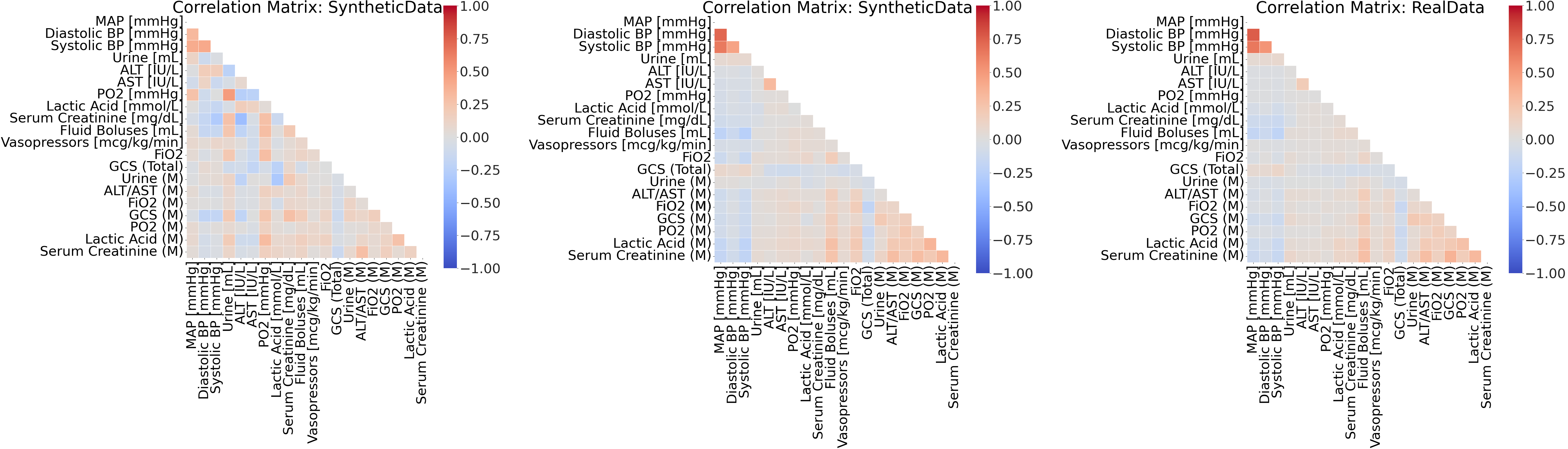}
      \caption{The dynamic correlations in trends.}
    \end{subfigure}

    \begin{subfigure}{\linewidth}
      \centering
      \includegraphics[width=\linewidth]{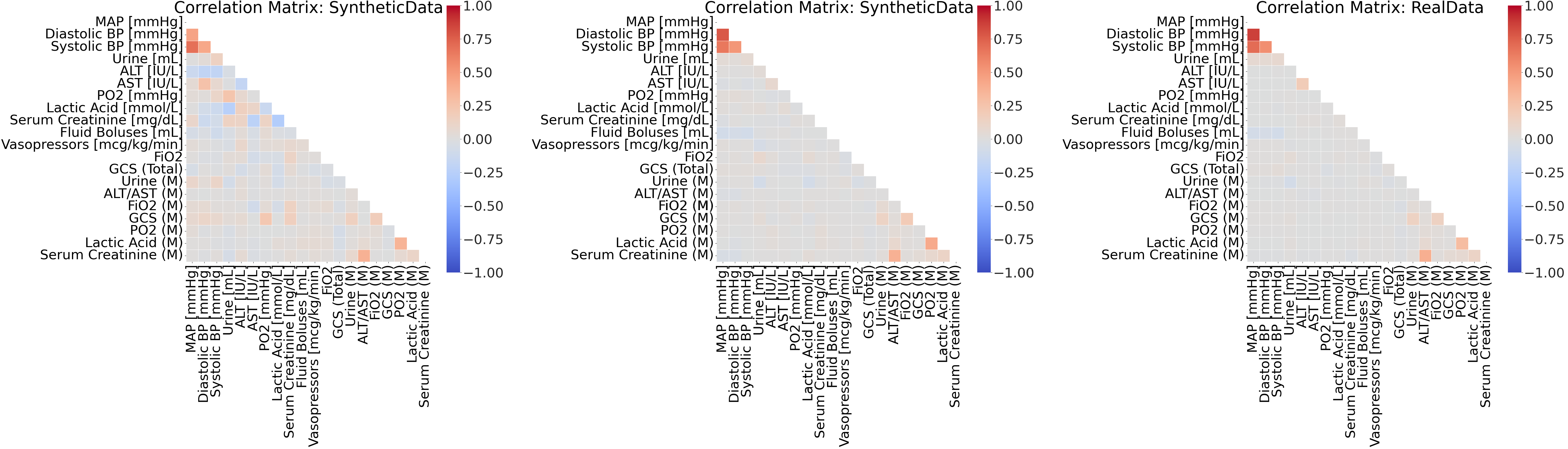}
      \caption{The dynamic correlations in cycles.}
    \end{subfigure}
    
    \caption{\label{Fig:CorrHypotention}Comparing the correlations in acute hypotension.\\
    The left panels depicts correlations in \citet{kuo2022health}'s $\mathfrak{D}_\text{null}$. Whereas the middle and right panels respectively depict the correlations in our $\mathfrak{D}_\text{alt}$ and those in the ground truth $\mathfrak{D}_\text{real}$.}
\end{figure}

\subsection{On the Correlations of the Variables}\label{Sec:CrossFiel}
\textbf{Acute Hypotension}\\
The correlations for acute hypotension are depicted in Figure \ref{Fig:CorrHypotention}. All panels on the left correspond to the synthetic dataset $\mathfrak{D}_\text{null}$ generated by \citet{kuo2022health}'s GAN; the middle panels represent our DPM-simulated dataset $\mathfrak{D}_\text{alt}$; and all panels on the right correspond to the ground truth dataset $\mathfrak{D}_\text{real}$. Furthermore, Figure \ref{Fig:CorrHypotention}(a) shows the static correlations, Figure \ref{Fig:CorrHypotention}(b) illustrates the dynamic correlations in trends, and Figure \ref{Fig:CorrHypotention}(c) presents the dynamic correlations in cycles.

Figure \ref{Fig:CorrHypotention} indicates that the correlations in our DPM-simulated dataset (located in the middle panels) exhibit a stronger resemblance to their real counterparts (located in the right panels) than those generated by GAN (located in the left panels). This applied to all three types of correlations considered, including static correlations as well as two types of dynamic correlations.

\textbf{ART for HIV}\\
Refer to all results in Appendix \ref{App:Stats}. The correlations are shown in Figure \ref{Fig:CorrHIV}. Both \citet{kuo2022health}'s $\mathfrak{D}_\text{null}$ and our DPM-simulated $\mathfrak{D}_\text{alt}$ capture the ground truth correlations. However, we noted that our dataset $\mathfrak{D}_\text{alt}$ exhibits a closer alignment with the real dataset $\mathfrak{D}_\text{real}$; while the correlations in $\mathfrak{D}_\text{null}$ tend to be exaggerated for both the static and dynamic correlations.

\newpage
\begin{table}[ht!]
    \centering
    \begin{tabular}{|l||l|l|r|}
        \hline
        \multicolumn{2}{|l|}{\textbf{Dataset}} & $U$ & \textbf{CAT}\\
        \hline
        \hline
        \multirow{2}{*}{Acute Hypotension} &  $\mathfrak{D}_\text{null}$~\citep{kuo2022health} & -2.1413 & 98.03\%\\
        \cline{2-4}
         &  $\mathfrak{D}_\text{alt}$ (ours) & \textbf{-2.4103} & \textbf{100.00}\%\\
        \hline
        \hline

        \multirow{2}{*}{ART for HIV} &  $\mathfrak{D}_\text{null}$~\citep{kuo2022health} & -2.130 & 97.50\%\\
        \cline{2-4}
         &  $\mathfrak{D}_\text{alt}$ (ours) & \textbf{-3.057} & \textbf{100.00}\%\\
        \hline
    \end{tabular}
    
    \caption{\label{Tab:LogUCover}A Comparison of the log-cluster metric ($U$) and the category coverage (CAT).\\
    It is the lower the better ($\downarrow$) for $U$; and higher the better ($\uparrow$) for CAT.}
\end{table}

\begin{figure}[ht!]
    \centering

      \includegraphics[width=0.8\linewidth]{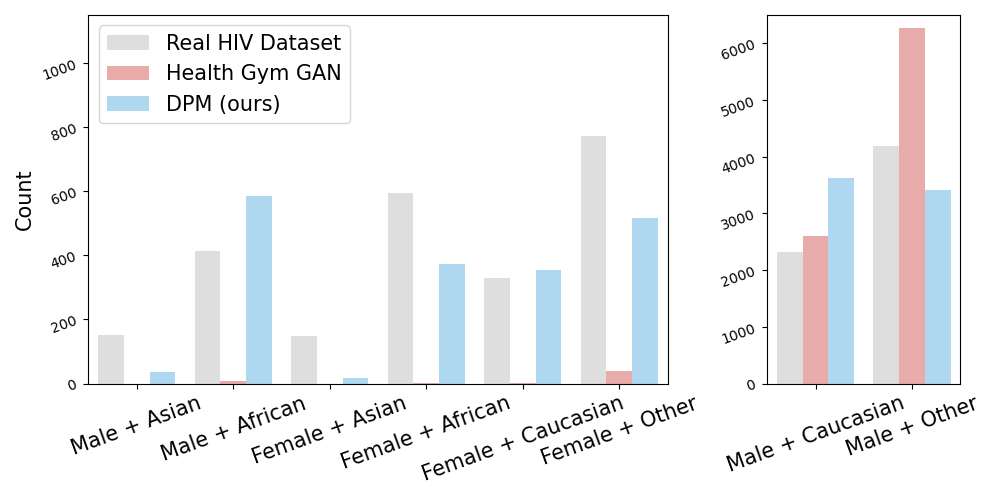}
    
    \caption{\label{Fig:Demo}Comparing the patient demographics in the ART for HIV datasets.\\
    We visualise the combination of \texttt{Gender} and \texttt{Ethnicity}. Colours grey, pink, and blue respectively indicate the ground truth $\mathfrak{D}_\text{real}$, \citet{kuo2022health}'s $\mathfrak{D}_\text{null}$, and our $\mathfrak{D}_\text{alt}$.}
\end{figure}

\subsection{On Diversity and Data Structure}

We calculated the log-cluster metric ($U$) and category coverage (CAT) to quantitatively assess the similarity between the latent structure of synthetic datasets and their real counterparts. The outcomes are summarised in Table \ref{Tab:LogUCover}. The CAT score showed that all combinations of binary and categorical variables are present in our DPM-simulated dataset $\mathfrak{D}_\text{alt}$; but such is not the case for not all combinations in the GAN-generated dataset $\mathfrak{D}_\text{null}$ produced by \citet{kuo2022health}. Moreover, the $U$ scores indicated that the latent structure embedded in our $\mathfrak{D}_\text{alt}$ is more realistic than that in $\mathfrak{D}_\text{null}$.

The metrics in Table \ref{Tab:LogUCover} can be more effectively contextualised through the qualitative analyses presented in Figure \ref{Fig:Demo}. For the patient demographics in the ART for HIV, we combined patient \texttt{Gender} and \texttt{Ethnicity}. The colours grey, pink, and blue correspond to the ground truth dataset $\mathfrak{D}_\text{real}$, \citet{kuo2022health}'s $\mathfrak{D}_\text{null}$, and our $\mathfrak{D}_\text{alt}$, respectively. Note, we conducted this analysis only for ART for HIV because the acute hypotension dataset does not include variables relating patient demographics.

We found that our $\mathfrak{D}_\text{alt}$ covers all combinations of demographic features present in the real dataset; whereas this is not the case for \citet{kuo2022health}'s $\mathfrak{D}_\text{null}$. This discrepancy suggests that mode collapse (as discussed in Section \ref{Sec:RelatedWork}) may be occurring in GANs and that while they can capture information relating distribution and correlation, synthetic data diversity is low and it remains challenging for GANs to accurately represent the complex, multi-faceted nature of clinical EHR data.

\subsection{Analysis of Risk Assessment Outcomes}

\textbf{Acute Hypotension}\\
The variables in the hypotension dataset (see Table \ref{Tab:Hypotension} in Appendix \ref{App:Variables}) are all related to the patient's bio-physiological states and do not contain any sensitive information that may reveal individuals' identities. Consequently, we only tested Euclidean distances and did not assess the disclosure risk. We found that records in our DPM-simulated synthetic dataset $\mathfrak{D}_\text{alt}$ matched none of those in the real hypotension dataset $\mathfrak{D}_\text{real}$. The minimum Euclidean distance between any synthetic record and any real record was 2.79 ($>0$), indicating that no data was leaked into the synthetic dataset.

\textbf{ART for HIV}\\
Likewise, the minimum Euclidean distance between any real and synthetic HIV record was 0.09 ($>0$), thus no real records was leaked into our synthetic dataset $\mathfrak{D}_\text{alt}$. The HIV variables (see Table \ref{Tab:HIV} in Appendix \ref{App:Variables}) contain the quasi-identifiers of \texttt{Gender} and \texttt{Ethnicity}. We combined these two variables to form distinct equivalence classes (\textit{e.g.,} male Asians and female Caucasians). The risk of a successful synthetic-to-real attack was estimated to be 0.076\%. This risk is also much lower than the standard threshold of 9\% (see Section \ref{Sec:SecEst}), signifying that our synthetic HIV dataset $\mathfrak{D}_\text{alt}$ can be released with minimal risk of sensitive information disclosure.

\subsection{Validation of Synthetic Dataset Utility}

\textbf{Acute Hypotension}\\
After training RL agents to suggest clinical treatments, we used heatmaps to visualize their action patterns. Each tile on the heatmap represents a unique action and its associated number indicates the frequency of that action as a proportion of all actions taken. 

\begin{figure}[ht!]
    \centering
    \begin{subfigure}{\linewidth}
      \centering
      \includegraphics[width=0.6\linewidth]{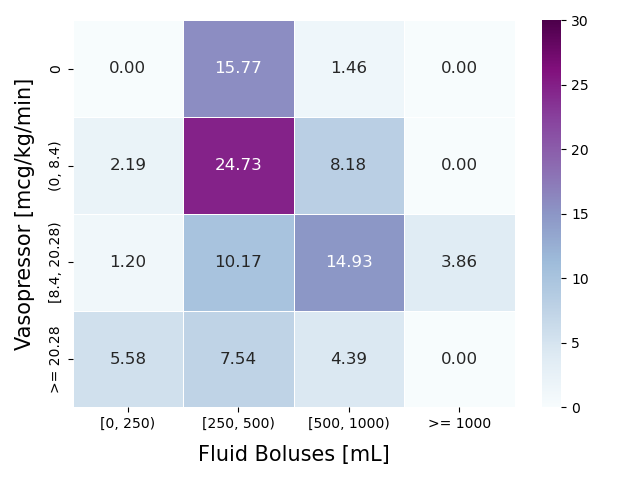}
      \caption{RL policy trained on the real dataset $\mathfrak{D}_\text{real}$.}
    \end{subfigure}
    
    \begin{subfigure}{\linewidth}
      \centering
      \includegraphics[width=0.6\linewidth]{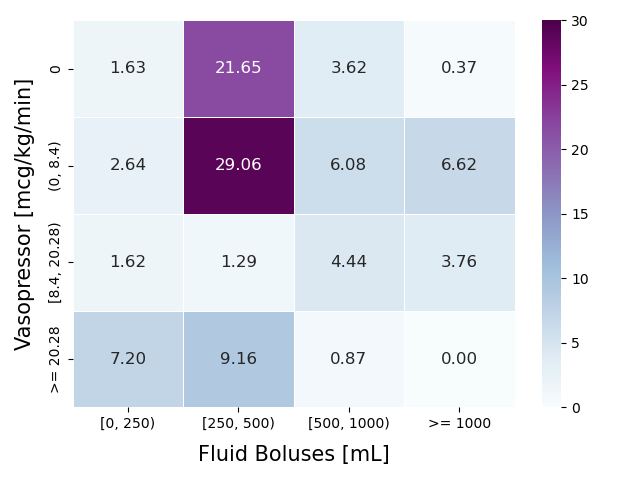}
      \caption{RL policy trained on \citet{kuo2022health}'s GAN-generated $\mathfrak{D}_\text{null}$.}
    \end{subfigure}

    \begin{subfigure}{\linewidth}
      \centering
      \includegraphics[width=0.6\linewidth]{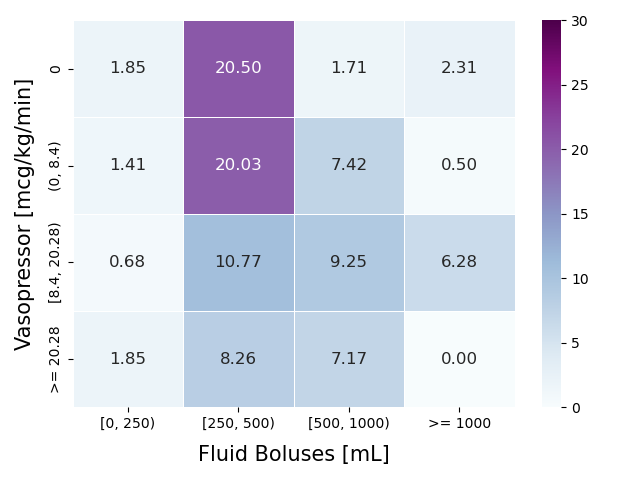}
      \caption{RL policy trained on our DPM-simulated $\mathfrak{D}_\text{alt}$.}
    \end{subfigure}
    
    \caption{\label{Fig:RlHypotension}Comparing the policies learned by RL agents on the acute hypotension datasets.\\
    We illustrate the recommended policies of RL agents, trained using various acute hypotension datasets. The RL action space is spanned by \texttt{Vasopressors} and \texttt{Fluid Boluses}.}
\end{figure}

We depicted the action patterns of the RL agents for acute hypotension in Figure \ref{Fig:RlHypotension}. The action space is spanned by \texttt{Vasopressor} and \texttt{Fluid Boluses}. Subplot (a) exhibits the actions taken by an RL agent trained on the real dataset $\mathfrak{D}_\text{real}$; subplots (b) and (c) respectively display the actions taken by RL agents trained on \citet{kuo2022health}'s synthetic dataset $\mathfrak{D}_\text{null}$ and our DPM-simulated $\mathfrak{D}_\text{alt}$. The heatmap in subplot (c) shows a better alignment with its counterpart in subplot (a), indicating that the RL agent trained on our $\mathfrak{D}_\text{alt}$ suggested actions that were more similar to those suggested by the RL agent trained on $\mathfrak{D}_\text{real}$.

\textbf{ART for HIV}\\
Refer to all results in Appendix \ref{App:Stats}. For ART for HIV, we also found that our DPM-simulated dataset for $\mathfrak{D}_\text{alt}$ has a higher utility than the GAN-generated $\mathfrak{D}_\text{null}$ by \citet{kuo2022health}. Figure \ref{Fig:RlHIV} shows that when the action space is spanned by \texttt{Comp. NNRTI} and \texttt{Base Drug Combo}, the RL agent trained on $\mathfrak{D}_\text{null}$ suggested the treatment of (NVP, DRV + FTC + TDF) for 48.97\% of all actions. This suggests that the GAN model used to generate $\mathfrak{D}_\text{null}$ experienced mode collapse, thus creating an excessive number of synthetic records with similar attributes in $\mathfrak{D}_\text{null}$. Conversely, we attribute the higher utility in the DPM-simulated $\mathfrak{D}_\text{alt}$ to the higher robustness of DPM against mode collapse.

\section{Caveats and Negative Outcomes}
In this section, we detail some negative outcomes and caveats of our DPM. While thus far we have demonstrated the model's ability to generate realistic synthetic datasets with high utility that are safe for public use, we have encountered difficulties in representing numeric variables with extremely long tails. Specifically, in the ART for HIV dataset simulated by our DPM, \texttt{VL} failed the KS test (see Table \ref{Tab:HIVStats}), and this issue could be further compounded when multiple long-tailed numeric variables are present.

To further investigate this issue, we tested our DPM on generating variables for a sepsis dataset based on the work of \citet{parbhoo2017combining}. Although our DPM was capable of generating a reliable sepsis dataset that passed the three sigma rule test, we observed that almost all of the numeric variables with extremely long tails failed the KS test. Moreover, the synthetic sepsis variables usually failed either one or both the t-test and F-test, indicating that our DPM had the tendency to learn biases towards numeric variables with extremely long tails.

Despite these limitations, we found that the synthetic sepsis dataset generated using our DPM exhibits realistic correlations and high diversity. However, we also observed a low utility of this dataset, where an RL agent trained on our synthetic dataset was unable to capture the suggested actions learned by an RL agent trained on the ground truth dataset. This was especially evident when numeric variables with extremely long tails spanned the action space.

In light of these results, we acknowledge the current limitations of our DPM in generating numeric variables with extremely long tails. However, we are committed to addressing this technical issue and plan to update our manuscript once we have optimised our DPM for the sepsis dataset. Refer to Appendix \ref{App:Sepsis} for a complete discussion on the synthesising a sepsis dataset using our DPM. 

\newpage

\newpage
\section{Discussion}
This paper presents a novel approach for generating realistic EHR data using DPMs. While recent work on DPMs~\citep{zheng2022diffusion, yuan2023ehrdiff} has shown promise in generating EHR data, these synthetic datasets remain simplistic. Our contribution lies in demonstrating that DPMs can simulate longitudinal EHR data over mixed-type variables, enabling downstream machine learning algorithms to be developed for advanced applications such as RL that require dynamic time-related information. To the best of our knowledge, this represented the first use of DPMs for this purpose.

We evaluated our DPM on three datasets, acute hypotension, ART for HIV, and sepsis, and compared its performance to that of GANs, specifically \citet{kuo2022health}'s Health Gym GAN. Our results show that our DPM generated more realistic datasets than GANs for the acute hypotension and ART for HIV datasets. Of note, our DPM-simulated variables better represented the multi-modal nature of clinical variables and showed better correlation alignment with the real datasets. Furthermore, the RL agent trained on our DPM-simulated datasets closely mirrored the policy learned by an agent trained on the real dataset, indicating higher utility than GAN-simulated datasets.

However, our experiments on sepsis revealed limitations in representing numeric variables with extremely long tails, leading to biases and low utility in certain applications. Our DPM tended to fail the KS, t-test, and F-test for such variables, highlighting the need for further optimisation.

\textbf{Data Records}\\
Below, we provide details on the synthetic hypotension dataset and the synthetic HIV dataset, which are hosted on PhysioNet and FigShare, respectively. Both datasets are stored as comma separated value (CSV) files and are accessible through \url{https://healthgym.ai/}.

The synthetic hypotension dataset comprises 3,910 synthetic patients and contains 48 data points per patient representing time-series of 48 hours. In total, there are 187,680 records (rows) in the dataset, with 22 variables (columns). The first 20 variables are organised as listed in Table \ref{Tab:Hypotension}, and the remaining two variables contain the synthetic patient IDs and the hour in the time series. The dataset is 26.0 MB in size and is generated to be realistic while being safe for public access.

Similarly, the synthetic HIV dataset contains 8,916 synthetic patients and time-series of 60 months with 60 data points per patient. The dataset comprises 534,960 records (rows) in total, with 15 variables, the first 13 of which are listed in Table \ref{Tab:HIV}. As with the hypotension dataset, the synthetic patient IDs and the month in the time series are contained in the remaining two variables. The dataset is 40.5 MB in size and is also generated to be realistic while being safe for public access.

\textbf{Broader Impact}\\
While our proposed DPM yielded synthetic datasets that are realistic and privacy-preserving, it is important to note that they should not be na\"ively considered as substitutes for actual datasets. In particular, there is a potential concern that synthetic data may carry forward existing biases or introduce new ones. To mitigate this issue, it is crucial to carefully select the features and data sources used to train the model and to regularly monitor the output data for biases.

Furthermore, despite observing similar optimal policies when comparing our DPM-simulated synthetic datasets with the ground truth, there is still room for further improvements. This suggests that our DPM model, although effective in capturing the complexities of a longitudinal EHR dataset with mixed-type variables, may require further fine-tuning to fully realise its potential.

\newpage
\bibliography{iclr2023_conference}
\bibliographystyle{iclr2023_conference}

\section*{Author Contributions Statement}
Corresponding author: Nicholas I-Hsien Kuo (\texttt{n.kuo@unsw.edu.au})

\textbf{N.K.} and \textbf{S.B.} designed, implemented and validated the deep learning models used to generate the synthetic datasets. \textbf{L.J.} contributed to the design of the study and provided expertise regarding the risk of sensitive information disclosure. Furthermore, \textbf{N.K.} wrote the manuscript and \textbf{S.B.} and \textbf{N.K.} designed the study. All authors contributed to the interpretation of findings and manuscript revisions.

\section*{Competing Interests}
The authors declare no competing interests.

\section*{Acknowledgements}
This study benefited from data provided by EuResist Network EIDB; and this project has been funded by a Wellcome Trust Open Research Fund (reference number 219691/Z/19/Z).

\newpage
\appendix
\section*{Supplementary Materials}

\textbf{Nicholas I-Hsien Kuo}, \textbf{Louisa Jorm}, \textbf{Sebastiano Barbieri}\\
Centre for Big Data Research in Health, the University of New South Wales, Sydney, Australia\\
\textcolor{white}{*}\\
Corresponding author: Nicholas I-Hsien Kuo (\texttt{n.kuo@unsw.edu.au})

The following appendix provides additional details and supporting information for the paper ``Synthetic Health-related Longitudinal Data with Mixed-type Variables Generated using Diffusion Models''. In the main text, we propose a novel method for generating synthetic datasets that capture the mixed-type variables of longitudinal EHRs using DPMs; and we present additional details and extra experimental outcomes in the supplementary materials. 

\section{Variables of the Datasets}\label{App:Variables}

The first dataset, for the management of acute hypotension, contains various clinical variables such as blood pressure and laboratory results. The second dataset, for HIV includes various medication combinations. Complete details and variable lists are provided in Tables \ref{Tab:Hypotension} and \ref{Tab:HIV}.

For data extraction and the inclusion/exclusion criteria, we mainly followed the Supplementary Information provided by \citet{kuo2022health} in \url{https://www.nature.com/articles/s41597-022-01784-7}. Additional guidelines on data formatting can be found in \citeauthor{kuo2022health}'s repository \url{https://github.com/Nic5472K/ScientificData2021_HealthGym}

\section{The Statistical Outcomes}\label{App:Stats}
\textbf{The Statistical Hypothesis Tests}\\
We implemented a series of hierarchical statistical tests as described in \citet{kuo2022health} to assess the realism of our synthetic variables (see Figure \ref{Fig:HST}). The results of these tests are presented in Tables \ref{Tab:HypotensionStats} and \ref{Tab:HIVStats}. Our objective was to determine whether the statistics of those data from the synthetic dataset used to train a neural network would be considered to be highly similar to the real dataset during iterative batch training. To achieve this, we sampled a batch of synthetic and real data with a batch size of $32$ for a maximum of $100$ iterations (hence the denominators in the Table are $100$). We then performed the four statistical tests in Figure \ref{Fig:HST} along the variable dimension. For a full description of the algorithm, please refer to Appendix D.5 on page 44 of \citet{kuo2022health}.

\textbf{The KL Divergences of the Individual Variables}\\
As mentioned in Section \ref{Sec:IndReal}, we used the KL divergence to estimate the similarity between the synthetic and real data distributions for each variable individually. To achieve this, we calculated the KL divergence between the true distribution and the learned distributions according to Equation (\ref{Eq:KldDiscrete}) and presented the full statistics in Tables \ref{Tab:HypotensionKLD} and \ref{Tab:HIVKLD}.

\begin{figure}[ht!]
    \centering
    \begin{subfigure}{\linewidth}
      \centering
      \includegraphics[width=0.825\linewidth]{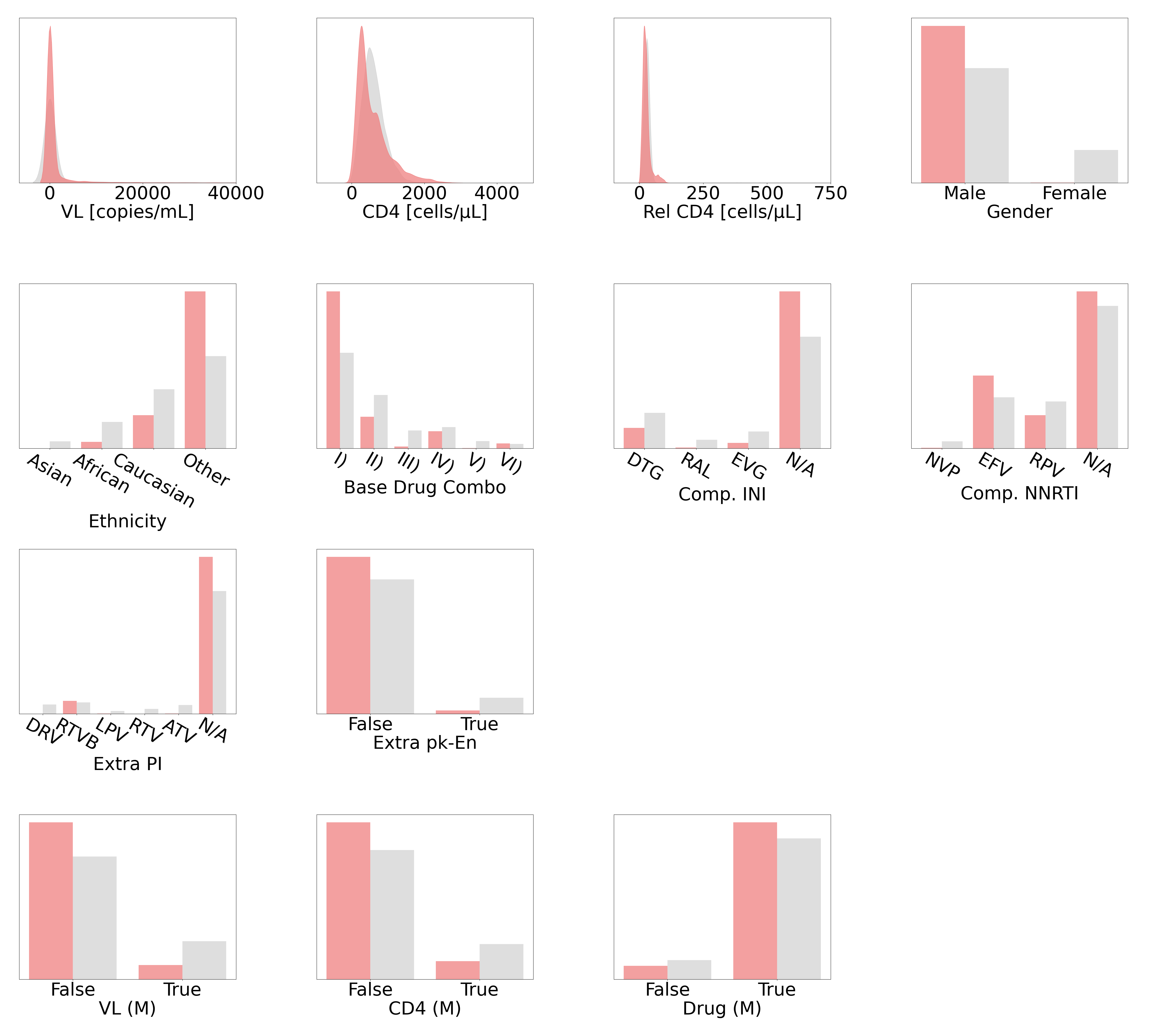}
      \caption{Synthetic dataset $\mathfrak{D}_\text{null}$ from \citet{kuo2022health} in pink.}
    \end{subfigure}
    
    \begin{subfigure}{\linewidth}
      \centering
      \includegraphics[width=0.825\linewidth]{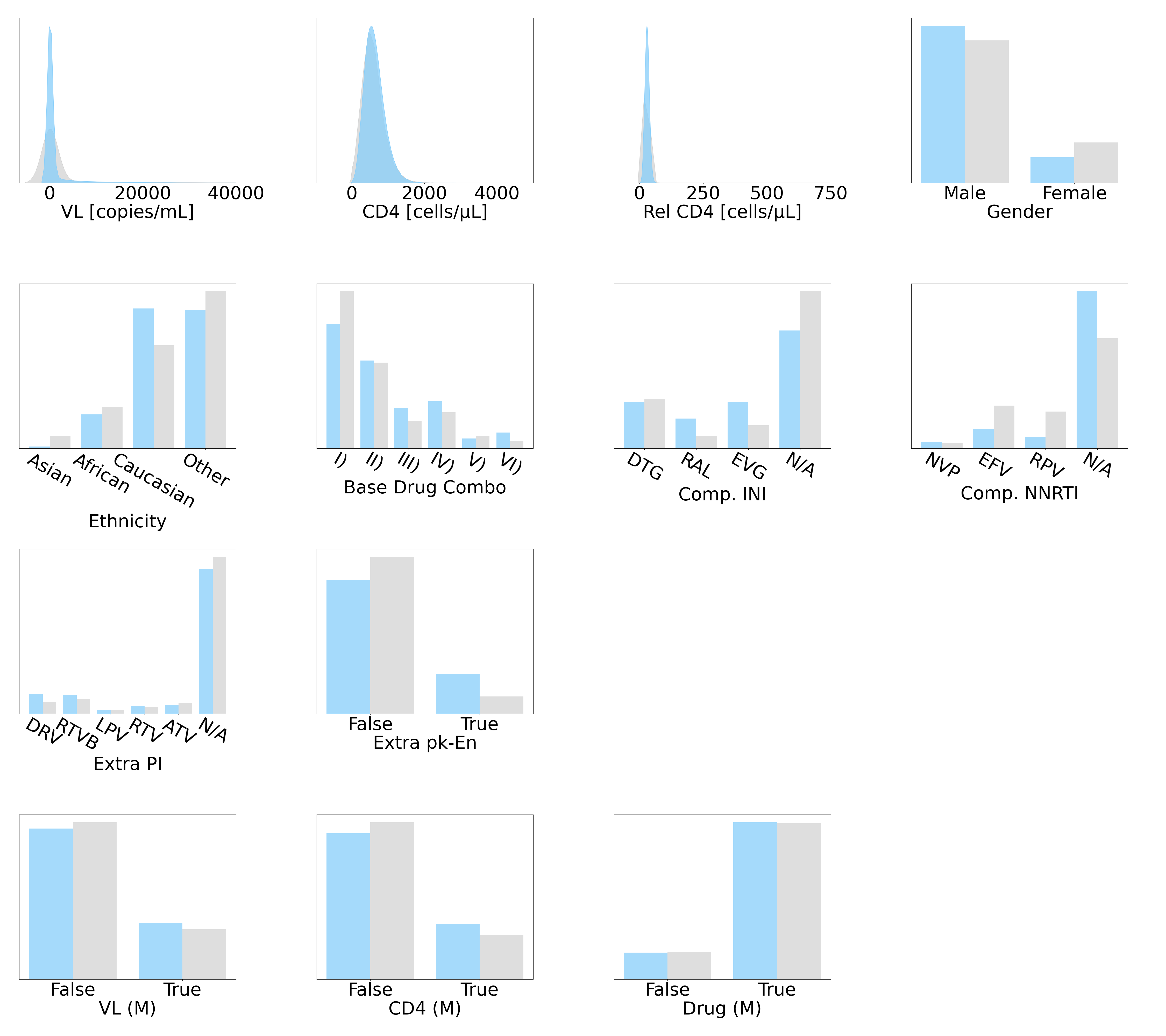}
      \caption{Synthetic dataset $\mathfrak{D}_\text{alt}$ from our DPM in blue.}
    \end{subfigure}%
    
    \caption{\label{Fig:KdeBarHIV}Comparing the variables in ART for HIV, with those of $\mathfrak{D}_\text{real}$ in colour grey.}
\end{figure}

\begin{figure}[ht!]
    \centering
    \begin{subfigure}{\linewidth}
      \centering
      \includegraphics[width=\linewidth]{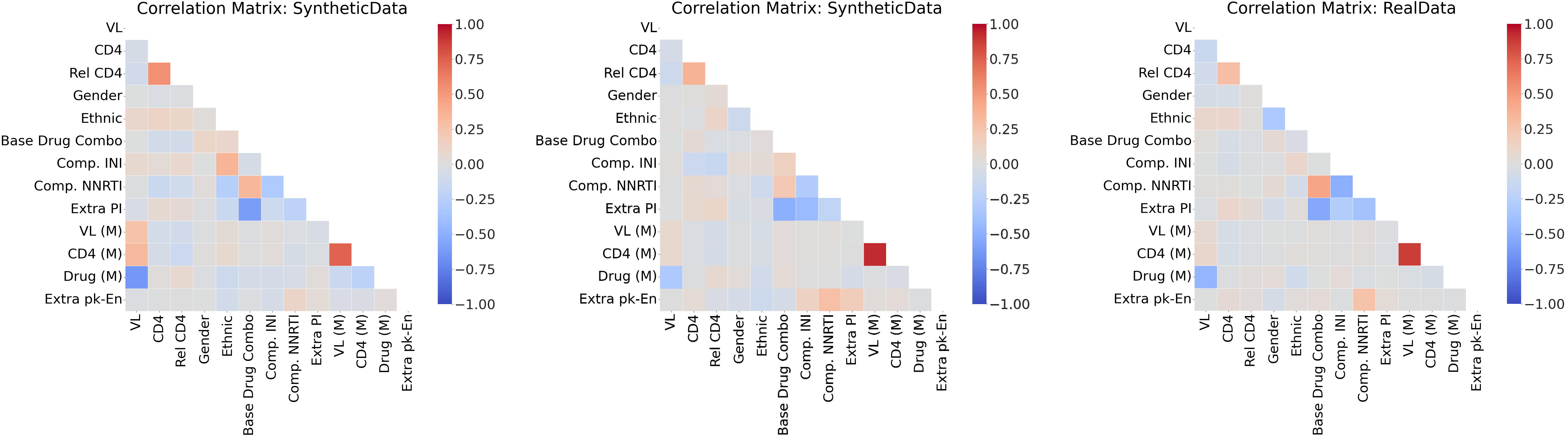}
      \caption{The static correlations.}
    \end{subfigure}
    
    \begin{subfigure}{\linewidth}
      \centering
      \includegraphics[width=\linewidth]{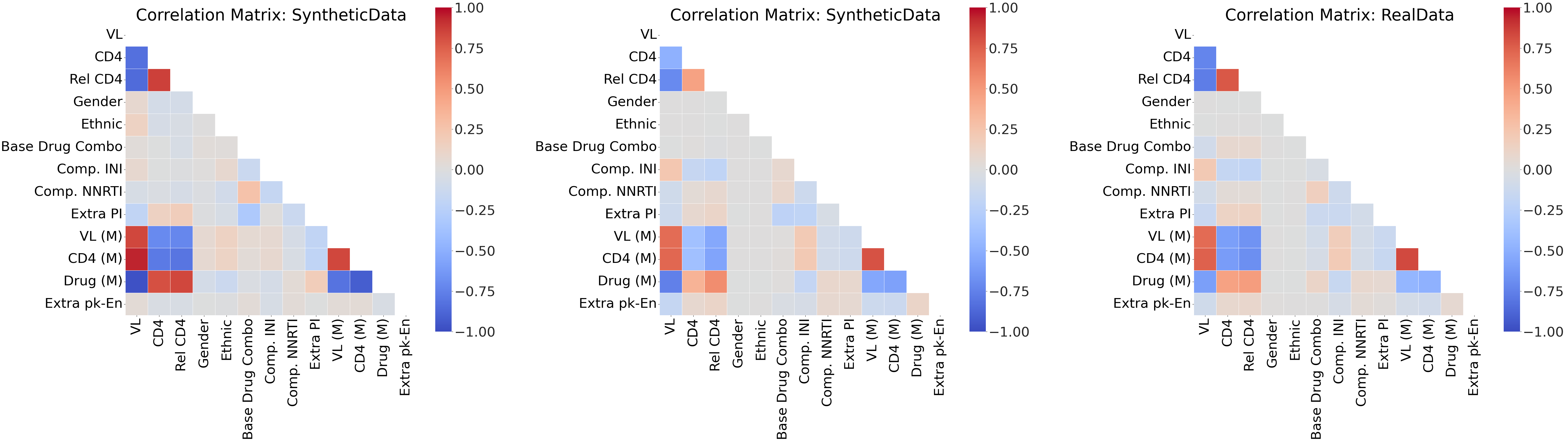}
      \caption{The dynamic correlations in trends.}
    \end{subfigure}

    \begin{subfigure}{\linewidth}
      \centering
      \includegraphics[width=\linewidth]{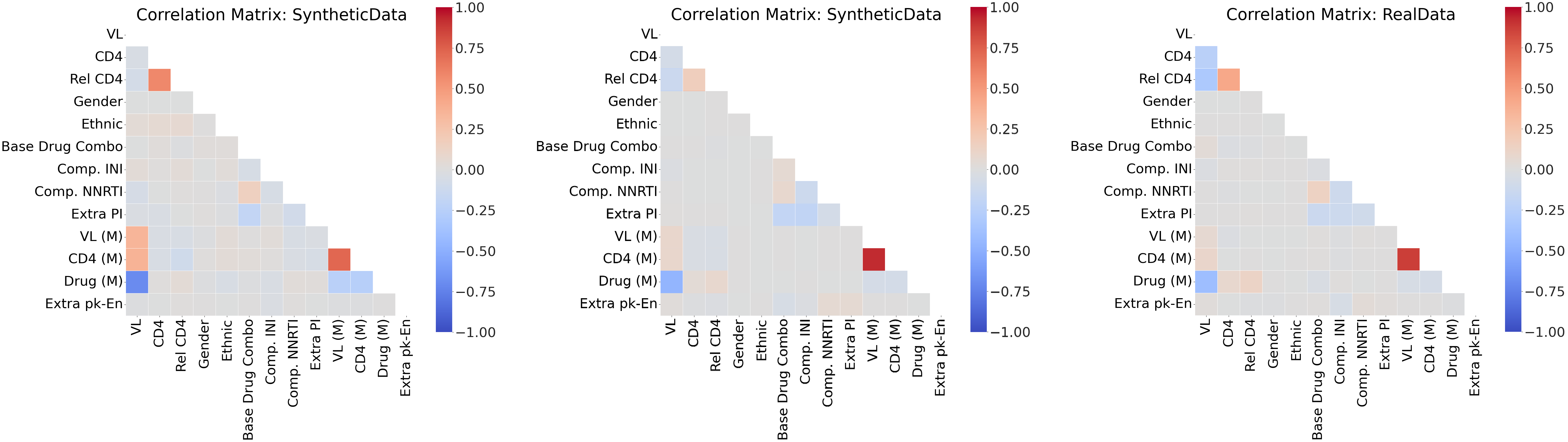}
      \caption{The dynamic correlations in cycles.}
    \end{subfigure}
    
    \caption{\label{Fig:CorrHIV}Comparing the correlations in ART for HIV.\\
    This figure is follows the configuration of Figure \ref{Fig:CorrHypotention}: the panels on the left, the middle, and the right represents the correlations from the synthetic dataset $\mathfrak{D}_\text{null}$ generated by \citet{kuo2022health}, from our DPM-simulated $\mathfrak{D}_\text{alt}$, and from the ground truth $\mathfrak{D}_\text{real}$, respectively.}
\end{figure}

\begin{figure}[ht!]
    \centering
    \begin{subfigure}{\linewidth}
      \centering
      \includegraphics[width=0.9\linewidth]{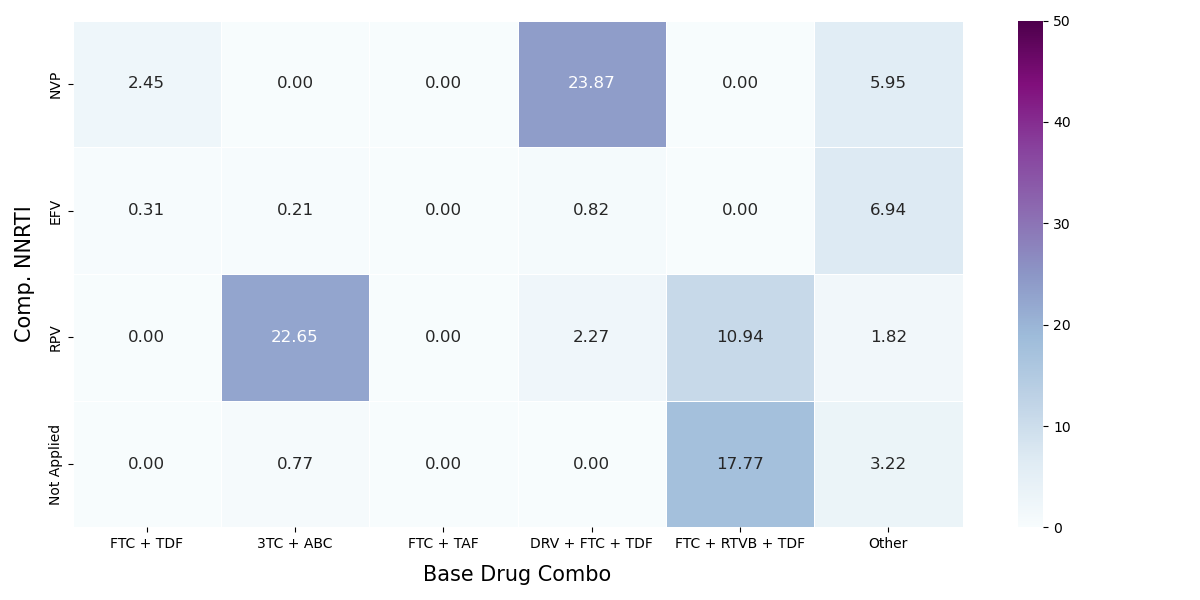}
      \caption{RL policy trained on the real dataset $\mathfrak{D}_\text{real}$.}
    \end{subfigure}
    
    \begin{subfigure}{\linewidth}
      \centering
      \includegraphics[width=0.9\linewidth]{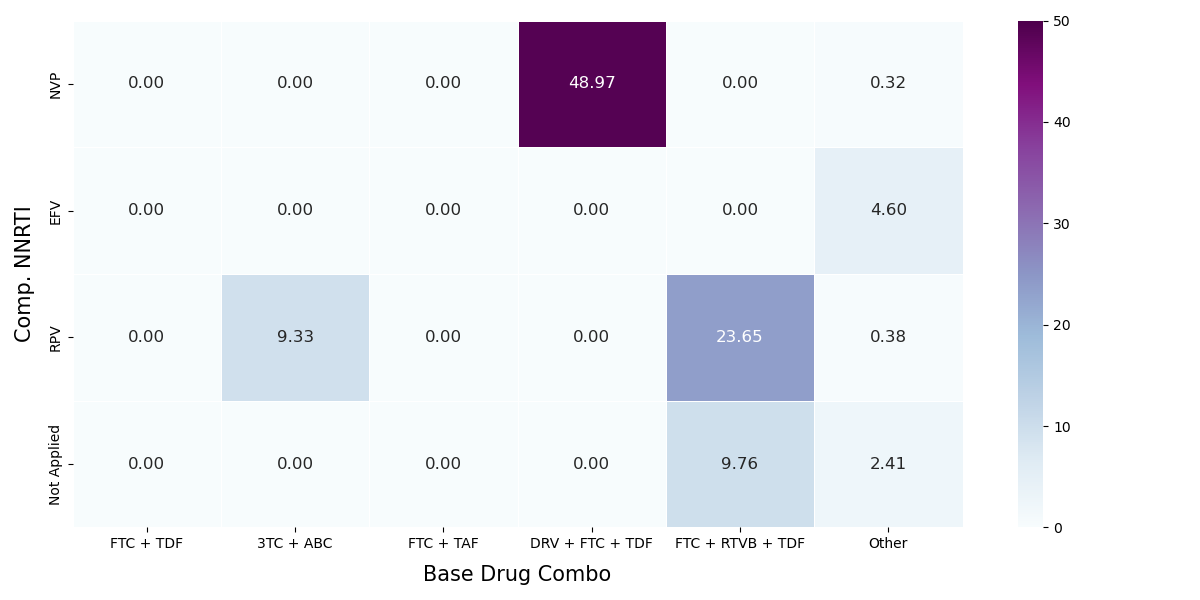}
      \caption{RL policy trained on \citet{kuo2022health}'s GAN-generated $\mathfrak{D}_\text{null}$.}
    \end{subfigure}

    \begin{subfigure}{\linewidth}
      \centering
      \includegraphics[width=0.9\linewidth]{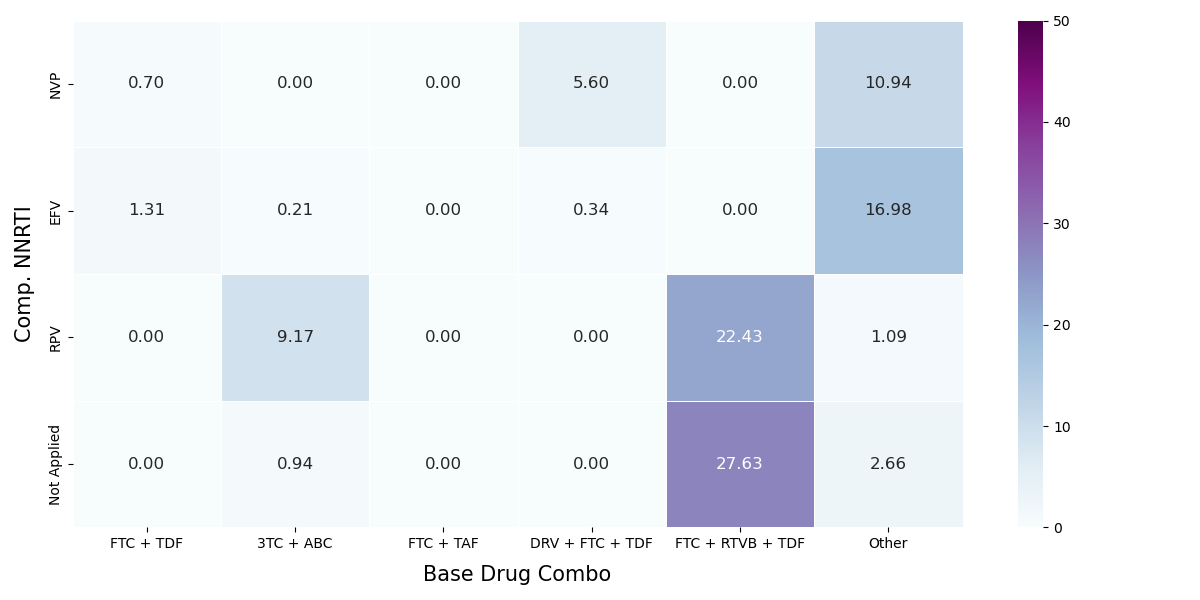}
      \caption{RL policy trained on our DPM-simulated $\mathfrak{D}_\text{alt}$.}
    \end{subfigure}
    
    \caption{\label{Fig:RlHIV}Comparing the policies learned by RL agents on the ART for HIV datasets.\\
    We illustrate the recommended policies of RL agents, trained using various ART for HIV datasets. The RL action space is spanned by \texttt{Comp. NNRTI} and \texttt{Base Drug Combo}.}
\end{figure}

\textbf{Extra Results on the ART for HIV}\\
To streamline our reporting of results, we have chosen to primarily focus on the findings concerning acute hypotension in the main text. In particular, we have relegated the additional outcomes pertaining to ART for HIV, which largely mirror the successes observed in the acute hypotension study, to the supplementary materials. Refer to the relevant outcomes illustrated in  Figures \ref{Fig:KdeBarHIV} -- \ref{Fig:RlHIV}.

\newpage
\begin{table}[ht]
    \footnotesize
    \centering
    \begin{tabular}{|l||l|l|l|}
        \hline
        \textbf{Variable Name} & 
        \textbf{Data Type} & \textbf{Unit} &
        \textbf{Extra Notes}\\
        
        \hline
        \hline
        Mean Arterial Pressure (MAP) & 
        \cellcolor{green!10}numeric & mmHg & \\
        
        \hline
        Diastolic Blood Pressure (DBP) & 
        \cellcolor{green!10}numeric & mmHg & \\

        \hline
        Systolic BP (SBP) & 
        \cellcolor{green!10}numeric & mmHg & \\

        \hline
        Urine & 
        \cellcolor{green!10}numeric & mL & \\

        \hline
        Alanine Aminotransferase (ALT) & 
        \cellcolor{green!10}numeric & IU/L & \\

        \hline
        Aspartate Aminotransferase (AST) & 
        \cellcolor{green!10}numeric & IU/L & \\

        \hline
        Partial Pressure of Oxygen (PaO$_2$) & 
        \cellcolor{green!10}numeric & mmHg & \\

        \hline
        Lactate & 
        \cellcolor{green!10}numeric & mmol/L & \\

        \hline
        Serum Creatinine &  
        \cellcolor{green!10}numeric & mg/dL & \\
        
        \hline
        \hline
        Fluid Boluses &  
        \cellcolor{brown!10}categorical & mL & {4 Classes}\\
        & & & 
        {$[0, 250)$; \hspace{3mm}\quad$[250, 500)$;}\\
        & & & 
        {$[500, 1000)$; \quad$\ge1000$}\\
        
        \hline
        Vasopressors &  
        \cellcolor{brown!10}categorical & mcg/kg/min & {4 Classes}\\
        & & & {$0$; \quad$(0, 8.4)$;}\\
        & & & {$[8.4, 20.28)$; \quad$\ge20.28$}\\
        
        \hline
        Fraction of Inspired Oxygen (FiO$_2$) &  
        \cellcolor{brown!10}categorical & fraction & {10 Classes}\\
        & & & {$\le0.2$; \quad$0.2$;}\\
        & & & {$0.3$; \quad$0.4$;}\\
        & & & {$0.5$; \quad$0.6$;}\\
        & & & {$0.7$; \quad$0.8$;}\\
        & & & {$0.9$; \quad$1.0$}\\
        
        \hline
        Glasgow Coma Scale Score (GCS) &  
        \cellcolor{brown!10}categorical & point & {13 Classes}\\
        & & & {$3$; \quad$4$; \quad$5$; \quad$6$;}\\
        & & & {$7$; \quad$8$; \quad$9$; \quad$10$;}\\
        & & & {$11$; \quad$12$; \quad$13$;}\\
        & & & {$14$; \quad$15$}\\
        
        \hline
        \hline
        Urine Data Measured (Urine (M)) &  
        \cellcolor{cyan!10}binary & - - & {False; \quad True}\\
        
        \hline
        ALT or AST Data Measured (ALT/AST (M)) &  
        \cellcolor{cyan!10}binary & - - & {False; \quad True}\\
        
        \hline
        FiO$_2$ (M) &  
        \cellcolor{cyan!10}binary & - - & {False; \quad True}\\
        
        \hline
        GCS (M) &  
        \cellcolor{cyan!10}binary & - - & {False; \quad True}\\
        
        \hline
        PaO$_2$ (M) &  
        \cellcolor{cyan!10}binary & - - & {False; \quad True}\\
        
        \hline
        Lactic Acid (M) &  
        \cellcolor{cyan!10}binary & - - & {False; \quad True}\\
        
        \hline
        Serum Creatinine (M) &  
        \cellcolor{cyan!10}binary & - - & {False; \quad True}\\

        \hline
    \end{tabular}
    
    \caption{\label{Tab:Hypotension}Variables in the Acute Hypotension Dataset\\
This table presents information related to the variables used in the acute hypotension datasets. In addition, we listed the different levels available to the non-numeric variables. 
}
\end{table}

\newpage
\begin{table}[ht]
    \footnotesize
    \centering
    \begin{tabular}{|l||l|l|l|}
        \hline
        \textbf{Variable Name} & 
        \textbf{Data Type} & \textbf{Unit} &
        \textbf{Extra Notes}\\
        
        \hline
        \hline
        Viral Load (VL) & 
        \cellcolor{green!10}numeric & copies/mL & \\
        
        \hline
        Absolute Count for CD4 (CD4) & 
        \cellcolor{green!10}numeric & cells/$\mu$L & \\

        \hline
        Relative Count for CD4 (Rel CD4) & 
        \cellcolor{green!10}numeric & cells/$\mu$L & \\
        
        \hline
        \hline
        Gender &  
        \cellcolor{cyan!10}binary & - - & \small{Female; \quad Male}\\
        
        \hline
        Ethnicity &  
        \cellcolor{brown!10}categorical & - - & \small{4 Classes}\\
        & & & \small{Asian; 
        \quad African;}\\
        & & & \small{Caucasian; 
        \quad Other}\\
        
        \hline
        \hline
        Base Drug Combination &  
        \cellcolor{brown!10}categorical & - - & \small{6 Classes}\\
        (Base Drug Combo) & & & 
        \small{FTC + TDF; 
        \quad 3TC + ABC;}\\
        & & & 
        \small{FTC + TAF; \quad DRV + FTC + TDF;}\\
        & & & 
        \small{FTC + RTVB + TDF; \quad Other}\\
        
        \hline
        Complementary INI &  
        \cellcolor{brown!10}categorical & - - & \small{4 Classes}\\
        (Comp. INI) & & & 
        \small{DTG; \quad RAL;}\\
        & & & 
        \small{EVG; \quad Not Applied}\\
        
        \hline
        Complementary NNRTI &  
        \cellcolor{brown!10}categorical & - - & \small{4 Classes}\\
        (Comp. NNRTI) & & & 
        \small{NVP; \quad EFV;}\\
        & & & 
        \small{RPV; \quad Not Applied}\\
        
        \hline
        Extra PI &  
        \cellcolor{brown!10}categorical & - - & \small{6 Classes}\\
        & & & 
        \small{DRV; \quad RTVB;}\\
        & & & 
        \small{LPV; \quad RTV;}\\
        & & & 
        \small{ATV; \quad Not Applied}\\
        
        \hline
        Extra pk Enhancer (Extra pk-En) &  
        \cellcolor{cyan!10}binary & - - & 
        \small{False; \quad True}\\
        
        \hline
        \hline
        VL Measured (VL (M)) &  
        \cellcolor{cyan!10}binary & - - & 
        \small{False; \quad True}\\
        
        \hline
        CD4 (M) &  
        \cellcolor{cyan!10}binary & - - & 
        \small{False; \quad True}\\
        
        \hline
        Drug Recorded (Drug (M)) &  
        \cellcolor{cyan!10}binary & - - & 
        \small{False; \quad True}\\

        \hline
    \end{tabular}
    
    \caption{\label{Tab:HIV}Variables in the ART for HIV Dataset\\
This table presents information related to the variables used in the ART for HIV datasets. In addition, we listed the different levels available to the non-numeric variables.
}
\end{table}

\newpage
\begin{table}[ht]
    \small
    \centering
    \begin{tabular}{|l||c|c|c|c|}
        \hline
        \textbf{Variable Name} & \textbf{\textcolor{white}{.}\hspace{3mm} KS-Test \textcolor{white}{.}\hspace{3mm}} & 
        \textbf{\textcolor{white}{.}\hspace{3mm} t-Test \textcolor{white}{.}\hspace{3mm}} & 
        \textbf{\textcolor{white}{.}\hspace{3mm} F-Test \textcolor{white}{.}\hspace{3mm}} & 
        \textbf{Three Sigma Rule Test}\\
        \hline
        \hline
        MAP & \cellcolor{cyan!10}$93/100$ & 
              \cellcolor{cyan!10}$90/100$ & 
              \cellcolor{cyan!10}$99/100$ & 
              \cellcolor{cyan!10}$100/100$\\
        \hline
        Diastolic BP & \cellcolor{cyan!10}$90/100$ & 
              \cellcolor{cyan!10}$86/100$ & 
              \cellcolor{cyan!10}$80/100$ & 
              \cellcolor{cyan!10}$100/100$\\
        \hline
        Systolic BP & \cellcolor{cyan!10}$93/100$ & 
              \cellcolor{cyan!10}$95/100$ & 
              \cellcolor{cyan!10}$91/100$ & 
              \cellcolor{cyan!10}$100/100$\\
        \hline
        Urine & \cellcolor{cyan!10}$88/100$ & 
                \cellcolor{cyan!10}$87/100$ & 
                \cellcolor{cyan!10}$98/100$ &
                \cellcolor{cyan!10}$99/100$\\
        \hline
        ALT & \cellcolor{magenta!10}$54/100$ & 
                \cellcolor{cyan!10}$91/100$ & 
                \cellcolor{cyan!10}$87/100$ &
                \cellcolor{cyan!10}$97/100$\\
        \hline
        AST & \cellcolor{magenta!10}$53/100$ & 
                \cellcolor{cyan!10}$91/100$ & 
                \cellcolor{cyan!10}$92/100$ &
                \cellcolor{cyan!10}$99/100$\\
        \hline
        PaO2 & \cellcolor{magenta!10}$46/100$ & 
                \cellcolor{cyan!10}$89/100$ & 
                \cellcolor{cyan!10}$76/100$ &
                \cellcolor{cyan!10}$99/100$\\
        \hline
        Lactic Acid & \cellcolor{magenta!10}$29/100$ & 
                \cellcolor{cyan!10}$85/100$ & 
                \cellcolor{cyan!10}$97/100$ &
                \cellcolor{cyan!10}$98/100$\\
        \hline
        Serum Creatinine & \cellcolor{cyan!10}$89/100$ & 
              \cellcolor{cyan!10}$83/100$ & 
              \cellcolor{cyan!10}$97/100$ & 
              \cellcolor{cyan!10}$100/100$\\
        \hline
        \hline
        Fluid Boluses & \cellcolor{cyan!10}$100/100$ & 
              \cellcolor{gray!10}- - & 
              \cellcolor{cyan!10}$87/100$ & 
              \cellcolor{gray!10}- -\\
        \hline
        Vasopressors & \cellcolor{cyan!10}$95/100$ & 
              \cellcolor{gray!10}- - & 
              \cellcolor{cyan!10}$89/100$ & 
              \cellcolor{gray!10}- -\\
        \hline
        FiO$_2$ & \cellcolor{cyan!10}$94/100$ & 
              \cellcolor{gray!10}- - & 
              \cellcolor{cyan!10}$95/100$ & 
              \cellcolor{gray!10}- -\\
        \hline
        GCS & \cellcolor{cyan!10}$93/100$ & 
              \cellcolor{gray!10}- - & 
              \cellcolor{cyan!10}$86/100$ & 
              \cellcolor{gray!10}- -\\
        \hline
        \hline
        Urine (M) & \cellcolor{cyan!10}$98/100$ & 
              \cellcolor{gray!10}- - & 
              \cellcolor{cyan!10}$95/100$ & 
              \cellcolor{gray!10}- -\\
        \hline
        ALT/AST (M) & \cellcolor{cyan!10}$100/100$ & 
              \cellcolor{gray!10}- - & 
              \cellcolor{cyan!10}$78/100$ & 
              \cellcolor{gray!10}- -\\
        \hline
        FiO$_2$ (M) & \cellcolor{cyan!10}$94/100$ & 
              \cellcolor{gray!10}- - & 
              \cellcolor{cyan!10}$95/100$ & 
              \cellcolor{gray!10}- -\\
        \hline
        GCS (M) & \cellcolor{cyan!10}$100/100$ & 
              \cellcolor{gray!10}- - & 
              \cellcolor{cyan!10}$98/100$ & 
              \cellcolor{gray!10}- -\\
        \hline
        PaO$_2$ (M) & \cellcolor{cyan!10}$100/100$ & 
              \cellcolor{gray!10}- - & 
              \cellcolor{cyan!10}$94/100$ & 
              \cellcolor{gray!10}- -\\
        \hline
        Lactic Acid (M) & \cellcolor{cyan!10}$100/100$ & 
              \cellcolor{gray!10}- - & 
              \cellcolor{cyan!10}$95/100$ & 
              \cellcolor{gray!10}- -\\
        \hline
        Serum Creatinine (M) & \cellcolor{cyan!10}$100/100$ & 
              \cellcolor{gray!10}- - & 
              \cellcolor{cyan!10}$96/100$ & 
              \cellcolor{gray!10}- -\\
        \hline

    \end{tabular}
    
    \caption{\label{Tab:HypotensionStats}Results on the hierarchical statistical tests for acute hypotension.}
\end{table}

\begin{table}[ht]
    \centering
    \begin{tabular}{|l||c|c|c|c|}
        \hline
        \textbf{Variable Name} & \textbf{\textcolor{white}{.}\hspace{3mm} KS-Test \textcolor{white}{.}\hspace{3mm}} & 
        \textbf{\textcolor{white}{.}\hspace{3mm} t-Test \textcolor{white}{.}\hspace{3mm}} & 
        \textbf{\textcolor{white}{.}\hspace{3mm} F-Test \textcolor{white}{.}\hspace{3mm}} & 
        \textbf{Three Sigma Rule Test}\\
        \hline
        \hline
        VL & 
        \cellcolor{magenta!10}$45/100$ & 
        \cellcolor{cyan!10}$94/100$ & 
        \cellcolor{magenta!10}$45/100$ & 
        \cellcolor{cyan!10}$100/100$\\
        \hline
        CD4 & 
        \cellcolor{cyan!10}$94/100$ & 
        \cellcolor{cyan!10}$92/100$ & 
        \cellcolor{cyan!10}$77/100$ & 
        \cellcolor{cyan!10}$99/100$\\
        \hline
        Rel CD4 & \cellcolor{cyan!10}$94/100$ & 
        \cellcolor{cyan!10}$87/100$ & 
        \cellcolor{cyan!10}$89/100$ & 
        \cellcolor{cyan!10}$100/100$\\
        \hline
        \hline
        Gender & 
        \cellcolor{cyan!10}$100/100$ & 
        \cellcolor{gray!10}- - & 
        \cellcolor{cyan!10}$100/100$ &
        \cellcolor{gray!10}- -\\
        \hline
        Ethnic & 
        \cellcolor{cyan!10}$99/100$ & 
        \cellcolor{gray!10}- - & 
        \cellcolor{cyan!10}$100/100$ &
        \cellcolor{gray!10}- -\\
        \hline
        \hline
        Base Drug Combo & \cellcolor{cyan!10}$97/100$ & 
        \cellcolor{gray!10}- - & 
        \cellcolor{cyan!10}$98/100$ &
        \cellcolor{gray!10}- -\\
        \hline
        Comp. INI & \cellcolor{cyan!10}$96/100$ & 
        \cellcolor{gray!10}- - & 
        \cellcolor{cyan!10}$100/100$ &
        \cellcolor{gray!10}- -\\
        \hline
        Comp. NNRTI & \cellcolor{cyan!10}$79/100$ & 
        \cellcolor{gray!10}- - & 
        \cellcolor{cyan!10}$97/100$ &
        \cellcolor{gray!10}- -\\
        \hline
        Extra PI & \cellcolor{cyan!10}$99/100$ & 
        \cellcolor{gray!10}- - & 
        \cellcolor{cyan!10}$100/100$ & 
        \cellcolor{gray!10}- -\\
        \hline
        Extra pk-En & \cellcolor{cyan!10}$99/100$ & 
        \cellcolor{gray!10}- - & 
        \cellcolor{cyan!10}$100/100$ & 
        \cellcolor{gray!10}- -\\
        \hline
        \hline
        VL (M) & 
        \cellcolor{cyan!10}$99/100$ & 
        \cellcolor{gray!10}- - & 
        \cellcolor{cyan!10}$100/100$ & 
        \cellcolor{gray!10}- -\\
        \hline
        CD4 (M) & \cellcolor{cyan!10}$100/100$ & 
        \cellcolor{gray!10}- - & 
        \cellcolor{cyan!10}$91/100$ & 
        \cellcolor{gray!10}- -\\
        \hline
        Drug (M) & \cellcolor{cyan!10}$99/100$ & 
        \cellcolor{gray!10}- - & 
        \cellcolor{cyan!10}$100/100$ & 
        \cellcolor{gray!10}- -\\
        \hline
        
    \end{tabular}
    
    \caption{\label{Tab:HIVStats}Results on the hierarchical statistical tests for ART for HIV.}
\end{table}

\newpage
\begin{table}[ht]
    \centering
    \begin{tabular}{|l||l|l|}
        \hline
        \textbf{Variable Name} & 
        \textbf{KL Divergence} from $\mathfrak{D}_\text{null}$ & 
        \textbf{KL Divergence} from $\mathfrak{D}_\text{alt}$ (Ours)\\
        \hline
        \hline
        MAP                 & 0.0724 &      0.0282\\
        \hline

        Diastolic BP        & 0.1837 &      0.0506\\
        \hline

        Systolic BP         & 0.1770 &      0.0374\\
        \hline

        Fluid Boluses       & 0.0014 &      0.1462\\
        \hline

        Urine               & 0.0155 &      0.0337\\
        \hline

        Vasopressors        & 0.0051 &      0.3881\\
        \hline

        ALT                 & 0.0196 &      0.0080\\
        \hline

        AST                 & 0.0203 &      0.0047\\
        \hline

        FiO$_2$             & 0.0123 &      0.0846\\
        \hline

        GCS                 & 0.0471 &      0.0469\\
        \hline

        PaO$_2$             & 0.2144 &      0.0929\\
        \hline

        Lactic Acid         & 0.0074 &      0.0842\\
        \hline

        Serum Creatinine    & 0.1270 &      0.0743\\
        \hline

        Urine (M)           & 0.0032 &      0.0076\\
        \hline

        ALT/AST (M)         & 9 $\times 10^{-5}$ & 1 $\times 10^{-5}$\\
        \hline

        FiO$_2$ (M)         & 0.0002 & 0.0068\\
        \hline

        GCS (M)             & 0.0047 & 0.0005\\
        \hline

        PaO$_2$ (M)         & 0.0002 & 0.0016\\
        \hline

        Lactic Acid (M)     & 1 $\times 10^{-5}$ & 0.0004\\
        \hline

        Serum Creatinine (M) & 0.0005 & 0.0004\\
        \hline

    \end{tabular}
    
    \caption{\label{Tab:HypotensionKLD}A comparison of the KL divergence of acute hypotension variables.}
\end{table}

\begin{table}[ht!]
    \centering
    \begin{tabular}{|l||l|l|}
        \hline
        \textbf{Variable Name} & 
        \textbf{KL Divergence} from $\mathfrak{D}_\text{null}$ & 
        \textbf{KL Divergence} from $\mathfrak{D}_\text{alt}$ (Ours)\\
        \hline
        \hline

        VL          & 0.3947 &      0.3482\\
        \hline

        CD4             & 0.0353 &      0.0318\\
        \hline

        Rel CD4             & 0.0080 &      0.0088\\
        \hline

        Gender         & 0.2230 &      0.0205\\
        \hline

        Ethnicity            & 0.2047 &      0.0473\\
        \hline

        Base Combo              & 0.1723 &      0.0360\\
        \hline

        INI                & 0.0947 &      0.0926\\
        \hline

        NNRTI                & 0.1921 &      0.1459\\
        \hline

        extra PI                 & 0.0597 &      0.0187\\
        \hline

        VL (M)                 & 0.0036 &      0.0022\\
        \hline

        CD4 (M)                & 0.0096 &      0.0078\\
        \hline

        Drug (M)                 & 0.0001 &      0.0001\\
        \hline

        pk-En                  & 0.0324 &      0.0725\\
        \hline
        
    \end{tabular}
    
    \caption{\label{Tab:HIVKLD}A comparison of the KLdivergence of ART for HIV variables.}
\end{table}

\newpage

\newpage

\newpage

\newpage
\section{Sepsis}\label{App:Sepsis}

We recognise that our DPM may not be completely fine-tuned or able to represent all EHR variables that exist. To increase transparency and openness in our research, we provide this appendix to outline the unfavorable outcomes of our DPM on a sepsis dataset. Specifically, we detail that their DPM is not yet optimised to represent numeric variables with particularly long tails.

\subsection{Background}
The real dataset for sepsis management, developed by \citet{komorowski2018artificial}, was extracted from the MIMIC-III database and encompasses 44 variables, including vital signs, laboratory results, mechanical ventilation information, and patient measurements. This dataset is comprised of time-series data for 2,164 patients, with varying durations of hospital stays ranging from 8 to 80 hours, with the data being reported in 4-hour windows, leading to 2 to 20 data points per patient record. The RL agent employed by \citeauthor{komorowski2018artificial} was trained to prescribe doses of intravenous fluids and vasopressors to patients, with rewards assigned based on patients transitioning to a more favorable health state following the actions taken. Refer to Tables \ref{Tab:SepsisNumeric} and \ref{Tab:SepsisNonNumeric} for more details.

\subsection{Experimental Setup}
The experimental setups used to simulate the sepsis dataset in this study closely follows the descriptions presented in Section \ref{Sec:DimChange}. However, certain modifications were implemented specifically for the sepsis data, which we discuss below.

Given the variable lengths of the sepsis data, zero-padding was utilised to bring all data to a pre-defined maximal length of $\mathfrak{L}_\text{sepsis} = 20$. Additionally, in line with the reasoning detailed in Section \ref{Sec:DimChange}, we determined $\mathfrak{N}_\text{sepsis} = 104$. Moreover, the sequence lengths for the sepsis data in the three resolution levels of the U-Net were 20, 5, and 3, respectively.

Training was conducted using a batch size of 32 and we set $T = 400$. The DPMs were trained until convergence, which was achieved after 2000 epochs.

To assess utility, we adopted \citet{raghu2017deep}'s reward function. See additional implementation details presented in Sections 7.1 and 7.2 of the Appendix of \citet{kuo2022health}.

\subsection{Individual Realism}\label{App:SepsisIndFiel}
\begin{figure}[ht!]
    \centering
    \begin{subfigure}{\linewidth}
      \centering
      \includegraphics[width=0.825\linewidth]{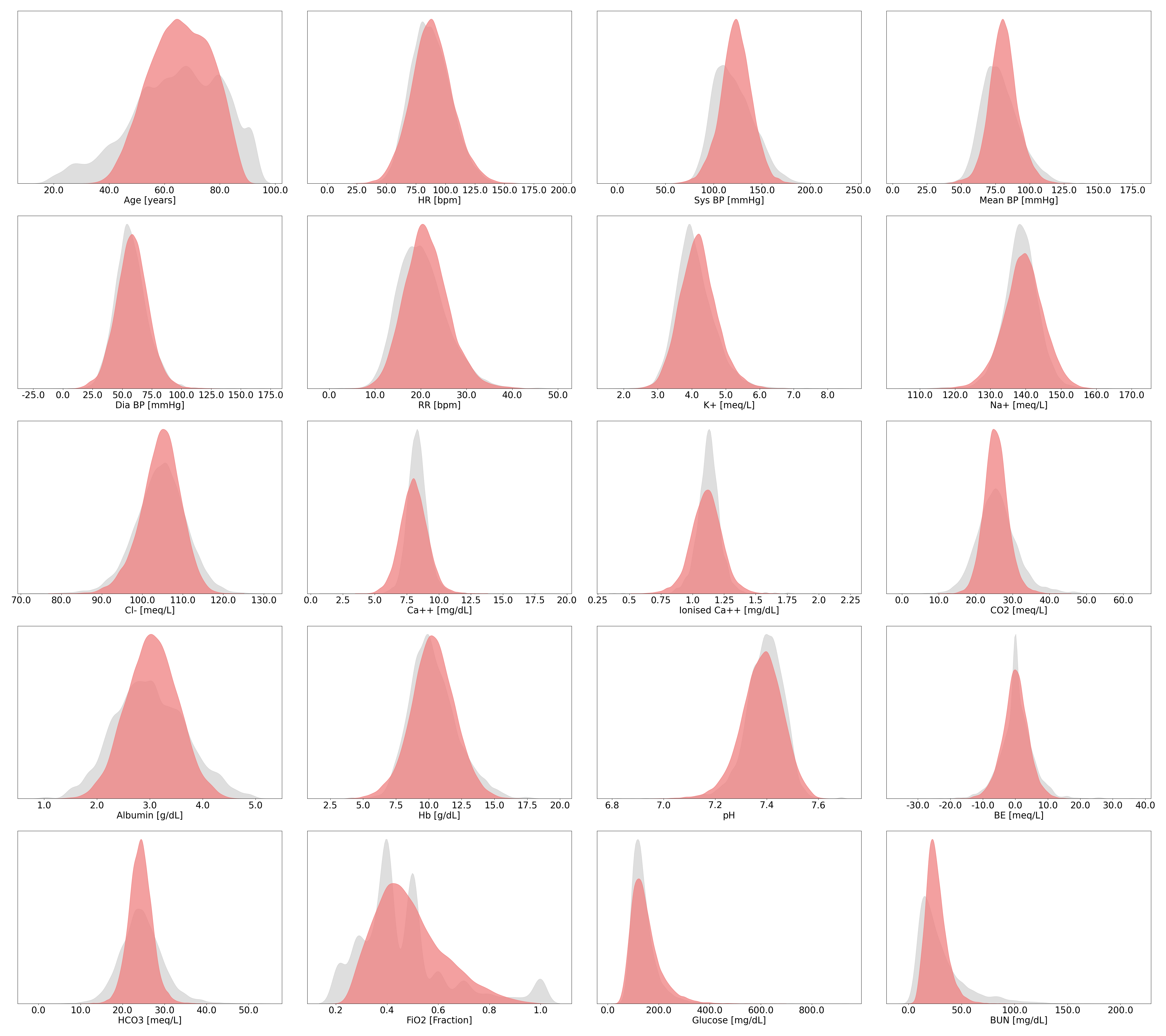}
      \caption{Synthetic dataset $\mathfrak{D}_\text{null}$ from \citet{kuo2022health} in pink.}
    \end{subfigure}
    \begin{subfigure}{\linewidth}
      \centering
      \includegraphics[width=0.825\linewidth]{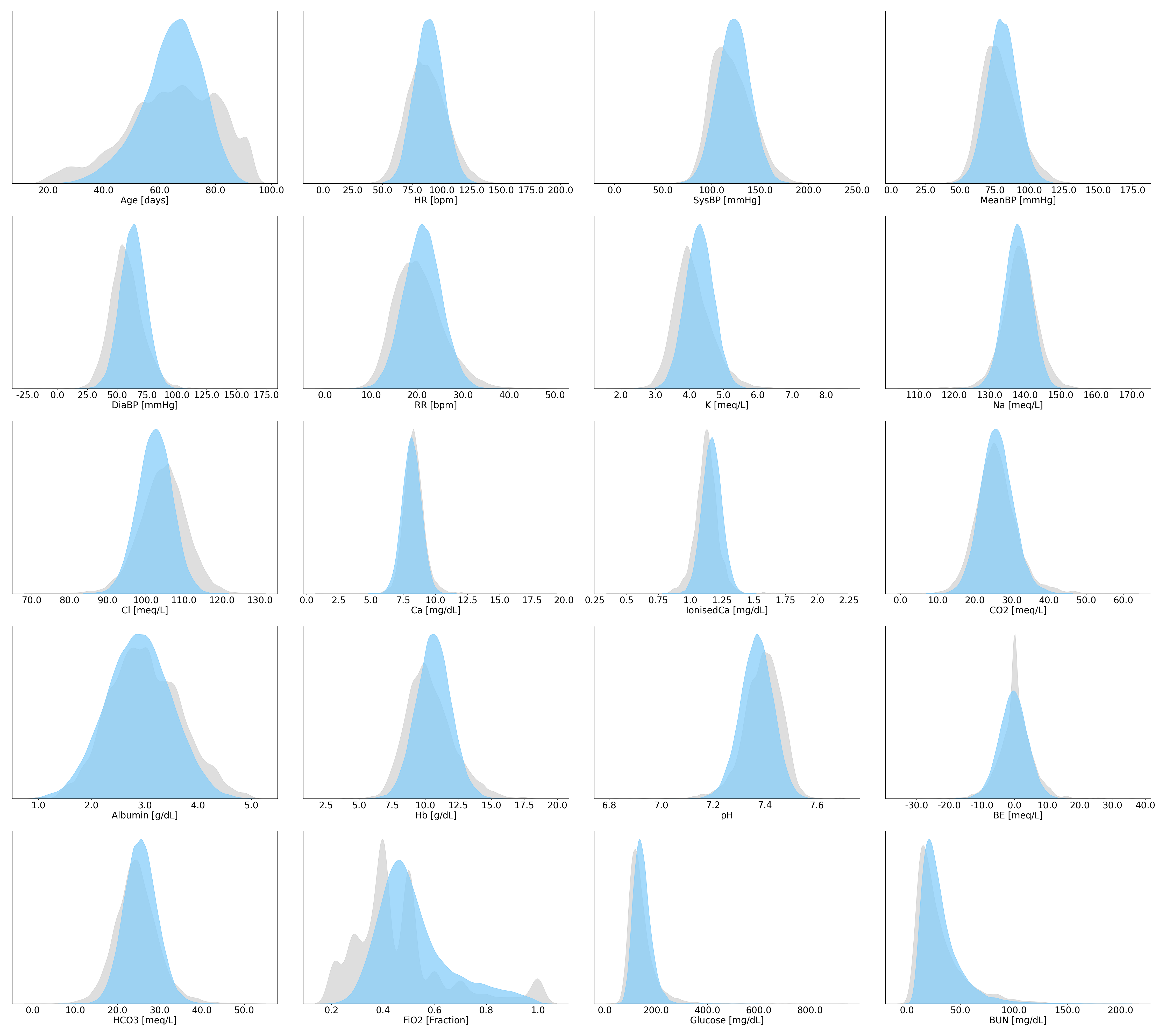}
      \caption{Synthetic dataset $\mathfrak{D}_\text{alt}$ from our DPM in blue.}
    \end{subfigure}
    \caption{\label{Fig:KdeBarSepsis01}Comparing one part of the variables in sepsis, with those of $\mathfrak{D}_\text{real}$ in colour grey.}
\end{figure}

\begin{figure}[ht!]
    \centering
    \includegraphics[width=\linewidth]{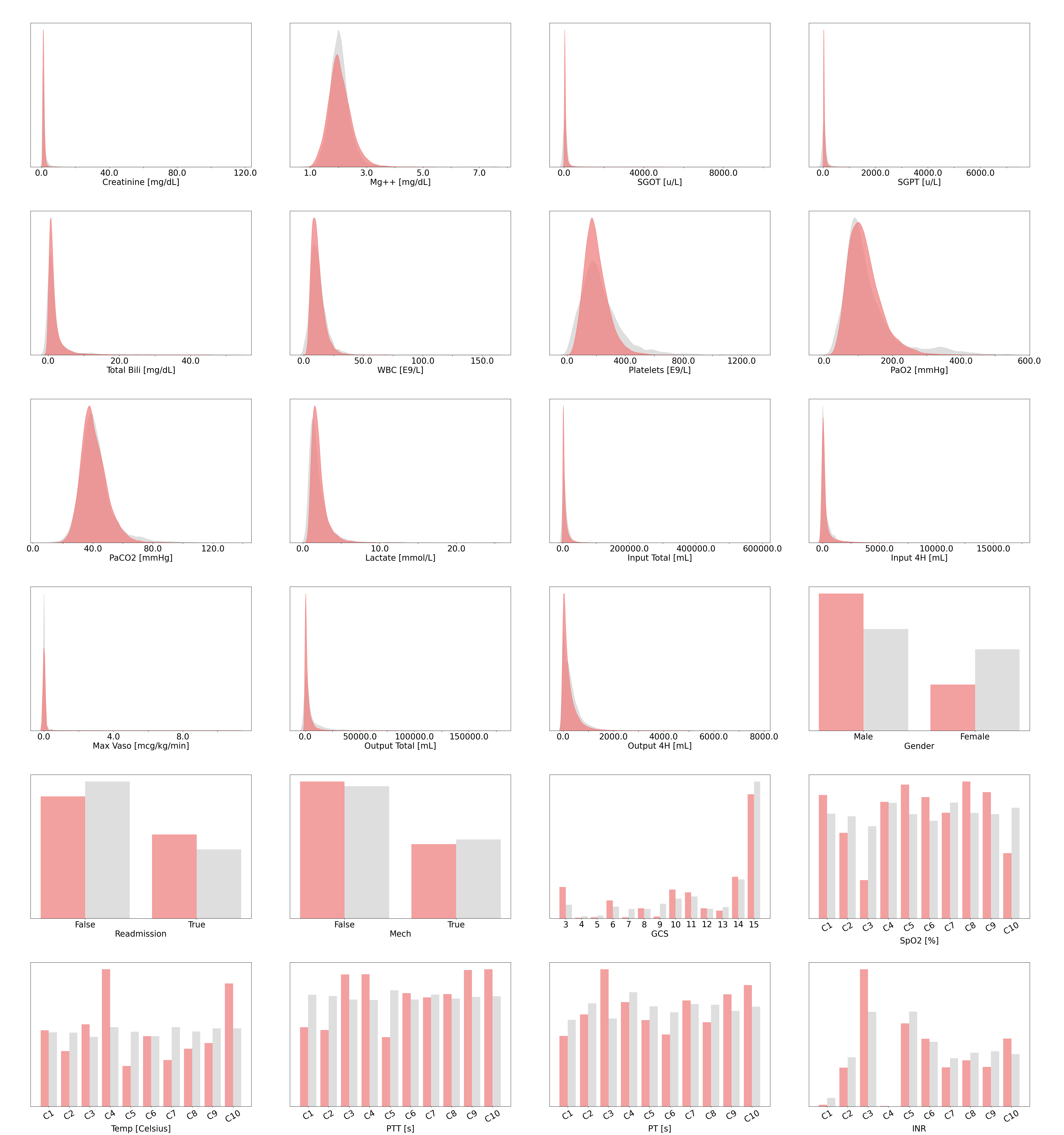}
    \caption{\label{Fig:KdeBarSepsis02}Additional the variables in sepsis.\\
    Those from the real datset $\mathfrak{D}_\text{real}$ are in colour grey and variables of the synthetic dataset $\mathfrak{D}_\text{null}$ generated by \citet{kuo2022health} are in pink.}
\end{figure}

The distributions of variables are depicted in Figures \ref{Fig:KdeBarSepsis01}, \ref{Fig:KdeBarSepsis02}, and \ref{Fig:KdeBarSepsis03}. The synthetic variables in $\mathfrak{D}_\text{null}$ from \citet{kuo2022health} and those in our $\mathfrak{D}_\text{alt}$ both effectively capture features of their genuine counterparts in $\mathfrak{D}_\text{real}$. To quantitatively assess the DPM-generated variables, we conducted statistical tests and present the results in Table \ref{Tab:SepsisStats}. The results showed that although all DPM-simulated variables are reliable, a noteworthy fraction of numeric variables (particularly those with significantly elongated tails) failed the KS test, which suggested that DPM may have the potential to encode biases in variable distributions. Notably, the KL divergences in Table \ref{Tab:SepsisKLD} revealed that the quality of DPM-simulated distributions is on-par with those generated using \citet{kuo2022health}'s GAN.

\subsection{Correlations}
The correlations of synthetic variables generated by \citet{kuo2022health}'s GAN and our DPM are compared with their real counterparts in Figure \ref{Fig:CorrSepsis}. Both \citet{kuo2022health}'s $\mathfrak{D}_\text{null}$ and our DPM-simulated $\mathfrak{D}_\text{alt}$ captured the ground truth correlations. Although a minor detail, it could be argued that $\mathfrak{D}_\text{null}$ better represented the dynamic correlations in trends, while $\mathfrak{D}_\text{alt}$ better represented the dynamic correlations in cycles.

\subsection{Data Diversity}

The results presented in Table \ref{Tab:LogUCover2} indicate that the latent structure in our DPM-simulated dataset is comparatively weaker. This finding can be attributed to the observations detailed in Appendix \ref{App:SepsisIndFiel}, wherein we identified the presence of numeric variables with extreme long tails in the sepsis dataset. Our DPM struggled to accurately capture the distributional features of such variables, leading to weaker latent structure in the simulated dataset.

\begin{figure}[ht!]
    \centering
    
      \includegraphics[width=0.8\linewidth]{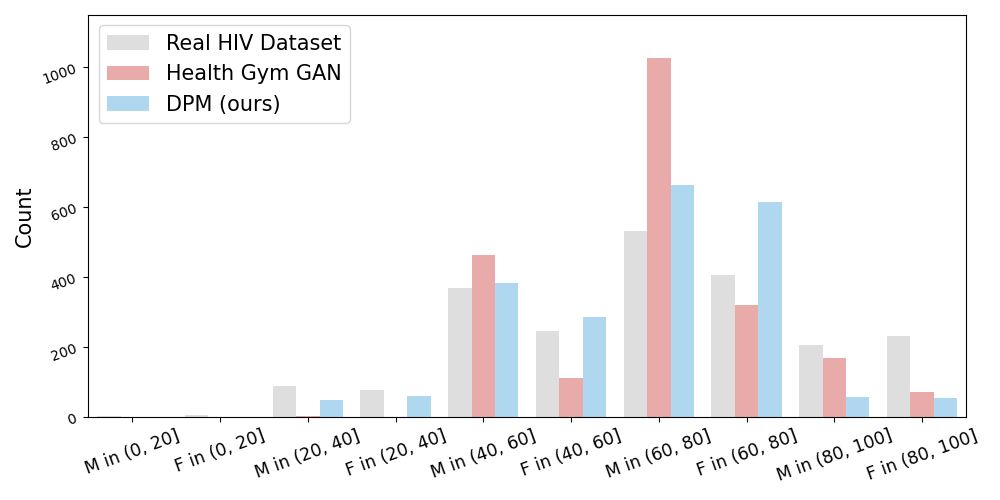}
    
    \caption{\label{Fig:Demo2}Comparing the patient demographics in the sepsis datasets.\\
    We selected the combination of \texttt{Gender} (M for male and F for female) and \texttt{Age} (grouped) for the sepsis cohort. The colours grey, pink, and blue respectively indicate the ground truth $\mathfrak{D}_\text{real}$, \citet{kuo2022health}'s GAN-generated $\mathfrak{D}_\text{null}$, and our DPM-simulated $\mathfrak{D}_\text{alt}$.}
\end{figure}

Nevertheless, when we examined data diversity in the context of Figure \ref{Fig:Demo2}, where patient demographics were compared across different cohorts using the \texttt{Age} variable binned into groups of 20-year intervals and combined with patient \texttt{Gender}, we discovered that our DPM-simulated datasets encompassed all possible combinations of demographic features present in the real dataset. This was not the case for \citet{kuo2022health}'s GAN-generated datasets. Hence, while DPMs may encode bias in numeric variables with extremely long tails, our results demonstrate that DPM-simulated data is better suited for representing cohort diversity. 

\subsection{Risk Assessment}
Using the Euclidean distance test, we discovered that none of the records in our synthetic dataset $\mathfrak{D}_\text{alt}$ were identical to any of the records in the real dataset $\mathfrak{D}_\text{real}$. The smallest distance between the real and synthetic records was 47.00 ($>0$).

This dataset also includes the quasi-identifiers of \texttt{Age} and \texttt{Gender}. We combined the \texttt{Age} variable (rounded down to the nearest year) and \texttt{Gender} to create different equivalence classes (\textit{e.g.,} males at 21 years old and females at 37 years old). The risk of a successful synthetic-to-real attack was estimated to be 5.44\%. This risk is significantly lower than the suggested threshold of 9\%, indicating that releasing our synthetic sepsis dataset $\mathfrak{D}_\text{alt}$ carries minimal risk of sensitive information disclosure.

\subsection{Utility}
\begin{figure}[ht!]
    \centering
    \includegraphics[width=\linewidth]{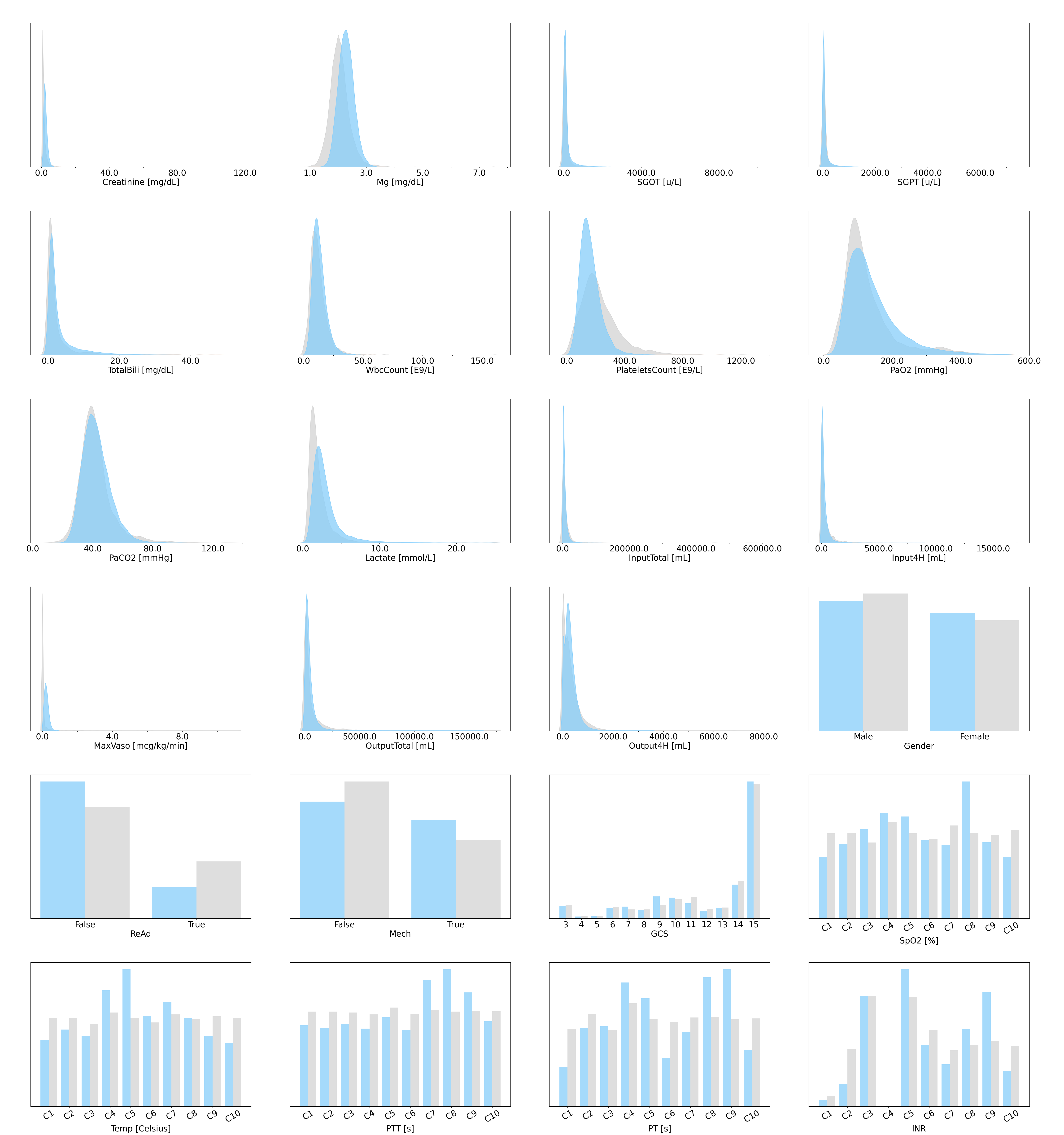}
    \caption{\label{Fig:KdeBarSepsis03}Additional the variables in sepsis.\\
    Those from the real datset $\mathfrak{D}_\text{real}$ are in colour grey and variables of the synthetic dataset $\mathfrak{D}_\text{alt}$ generated using our DPM are in blue.}
\end{figure}

\begin{figure}[ht!]
    \centering
    \begin{subfigure}{\linewidth}
      \centering
      \includegraphics[width=\linewidth]{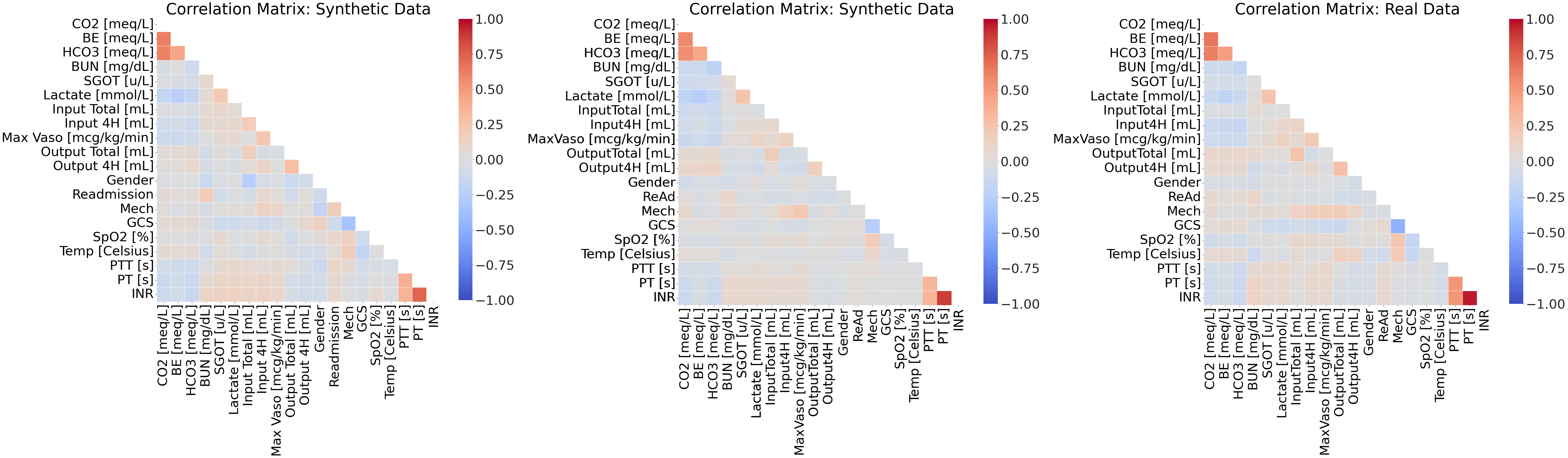}
      \caption{The static correlations.}
    \end{subfigure}
    
    \begin{subfigure}{\linewidth}
      \centering
      \includegraphics[width=\linewidth]{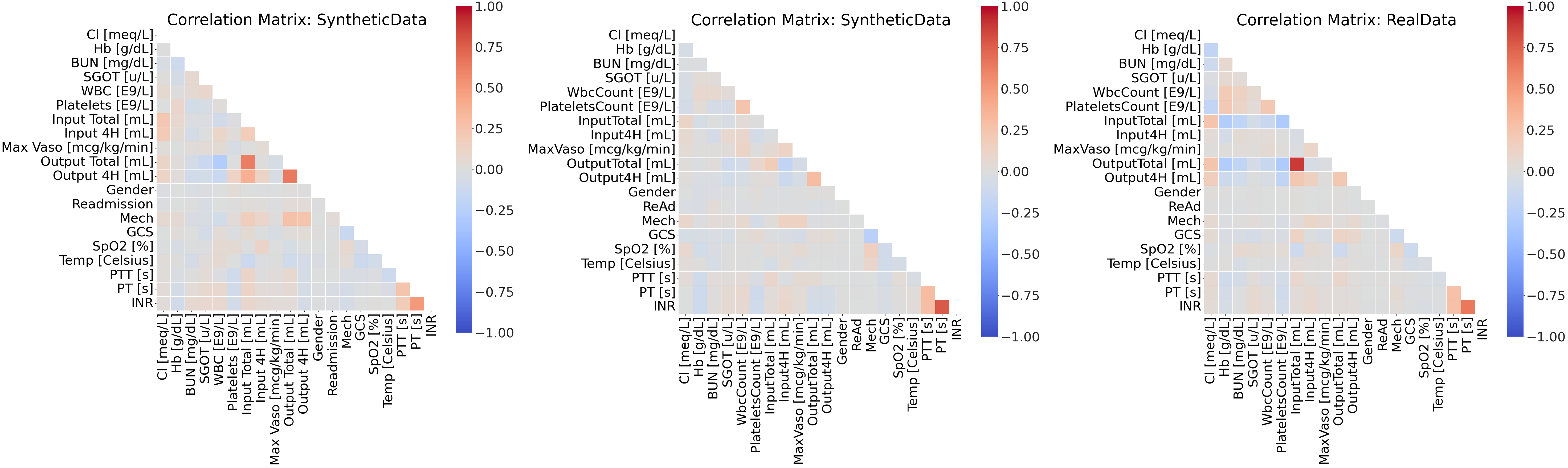}
      \caption{The dynamic correlations in trends.}
    \end{subfigure}

    \begin{subfigure}{\linewidth}
      \centering
      \includegraphics[width=\linewidth]{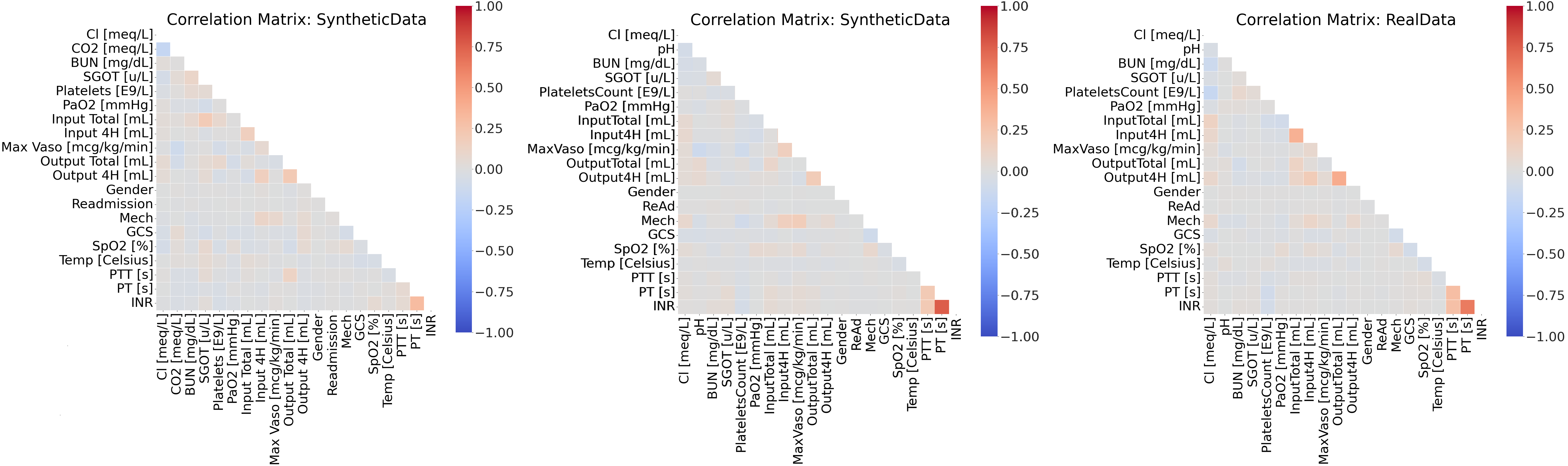}
      \caption{The dynamic correlations in cycles.}
    \end{subfigure}
    
    \caption{\label{Fig:CorrSepsis}Comparing the correlations in sepsis.\\
    This figure is follows the configuration of Figure \ref{Fig:CorrHypotention}: the panels on the left, the middle, and the right represents the correlations from the synthetic dataset $\mathfrak{D}_\text{null}$ generated by \citet{kuo2022health}, from our DPM-simulated $\mathfrak{D}_\text{alt}$, and from the ground truth $\mathfrak{D}_\text{real}$, respectively.}
\end{figure}

\begin{figure}[ht!]
    \centering
    \begin{subfigure}{\linewidth}
      \centering
      \includegraphics[width=0.6\linewidth]{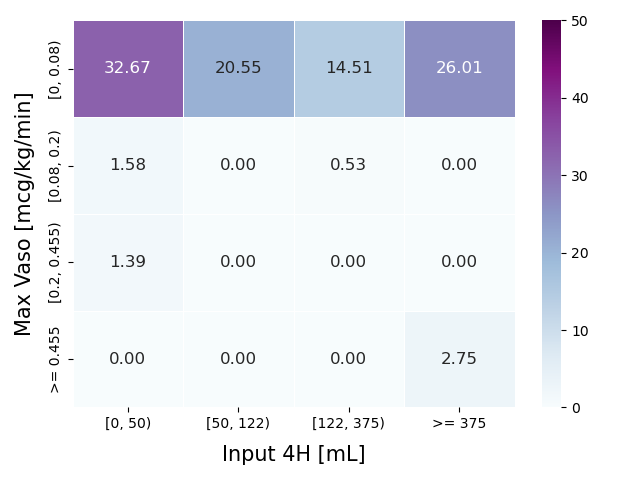}
      \caption{RL policy trained on the real dataset $\mathfrak{D}_\text{real}$.}
    \end{subfigure}
    
    \begin{subfigure}{\linewidth}
      \centering
      \includegraphics[width=0.6\linewidth]{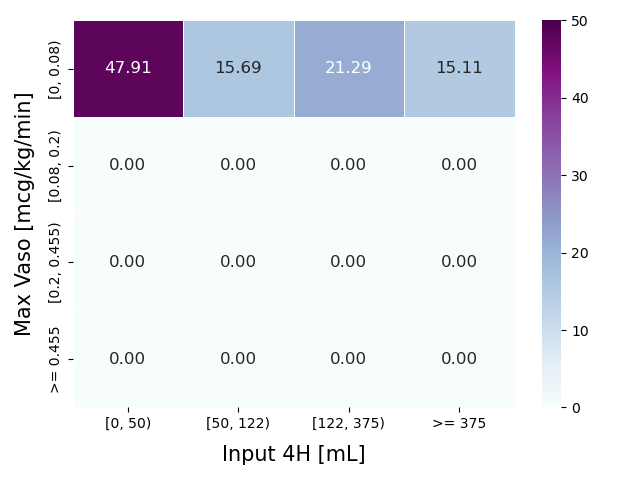}
      \caption{RL policy trained on \citet{kuo2022health}'s GAN-generated $\mathfrak{D}_\text{null}$.}
    \end{subfigure}

    \begin{subfigure}{\linewidth}
      \centering
      \includegraphics[width=0.6\linewidth]{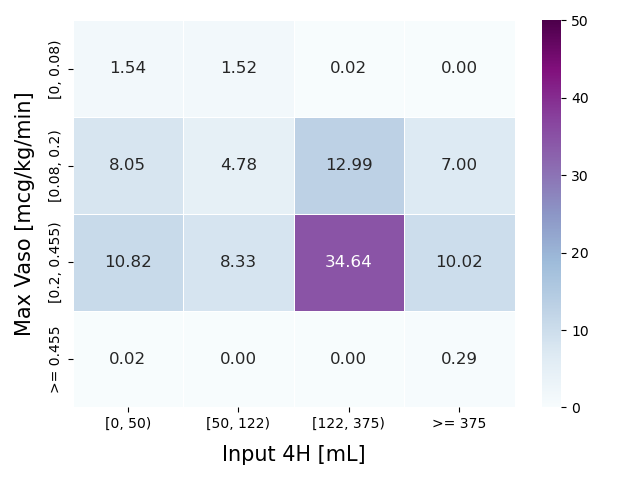}
      \caption{RL policy trained on our DPM-simulated $\mathfrak{D}_\text{alt}$.}
    \end{subfigure}
    
    \caption{\label{Fig:RlSepsis}Comparing the policies learned by RL agents on the sepsis datasets.\\
    We illustrate the recommended policies of RL agents, trained using various sepsis datasets. The RL action space is spanned by \texttt{Max Vaso} and \texttt{Input 4H}.}
\end{figure}

However, our experimentation revealed that the utility of our DPM-simulated $\mathfrak{D}_\text{alt}$ was unsatisfactory for the sepsis dataset. As depicted in Figure \ref{Fig:RlSepsis}, when the action space was spanned by \texttt{Max Vaso} and \texttt{Input 4H}, the RL agent trained on our $\mathfrak{D}_\text{alt}$ failed to replicate the actions taken by the RL agent trained on the ground truth $\mathfrak{D}_\text{real}$. This unfavorable outcome may have resulted from DPM's learning biases in the distribution of numeric values with extreme long tails, as discussed in Appendix \ref{App:SepsisIndFiel}. Although we previously demonstrated that the DPM-simulated sepsis dataset contained reliable variables (see Figure \ref{Fig:KdeBarSepsis03}), realistic correlations (see Figure \ref{Fig:CorrSepsis}), and higher diversity in patient demographics (see Figure \ref{Fig:Demo}), our $\mathfrak{D}_\text{alt}$ still fell short in achieving satisfactory utility.

\subsection{Concluding Remark}
Our DPM-simulated sepsis dataset did not meet the desired level of utility, and as such, we have chosen not to make it publicly available in its current state. We remain dedicated to addressing this limitation and will continue to explore potential solutions. As new findings emerge, we will update this manuscript to reflect our progress in this area.

\newpage
\begin{table}[ht]
    \footnotesize
    \centering
    \begin{tabular}{|l||l|l|l|}
        \hline
        \textbf{Variable Name} & 
        \textbf{Data Type} & \textbf{Unit} &
        \textbf{Extra Notes}\\
        
        \hline
        \hline
        Age & 
        \cellcolor{green!10}numeric & year & \\
        \hline
        
        Heart Rate (HR) & 
        \cellcolor{green!10}numeric & bpm & \\
        \hline
        
        Systolic BP & 
        \cellcolor{green!10}numeric & mmHg & \\
        \hline
        
        Mean BP & 
        \cellcolor{green!10}numeric & mmHg & \\
        \hline
        
        Diastolic BP & 
        \cellcolor{green!10}numeric & mmHg & \\
        \hline
        
        Respiratory Rate (RR) & 
        \cellcolor{green!10}numeric & bpm & \\
        \hline
        
        Potassium (K$^{+}$) & 
        \cellcolor{green!10}numeric & meq/L & \\
        \hline
 
        Sodium (Na$^{+}$) & 
        \cellcolor{green!10}numeric & meq/L & \\
        \hline
        
        Chloride (Cl$^{-}$) & 
        \cellcolor{green!10}numeric & meq/L & \\
        \hline
        
        Calcium (Ca) & 
        \cellcolor{green!10}numeric & mg/dL & \\
        \hline
        
        Ionised Ca$^{++}$ & 
        \cellcolor{green!10}numeric & mg/dL & \\
        \hline
        
        Carbon Dioxide (CO$_2$) & 
        \cellcolor{green!10}numeric & meq/L & \\
        \hline
        
        Albumin & 
        \cellcolor{green!10}numeric & g/dL & \\
        \hline
        
        Hemoglobin (Hb) & 
        \cellcolor{green!10}numeric & g/dL & \\
        \hline
        
        Potential of Hydrogen (pH) & 
        \cellcolor{green!10}numeric & - - & \\
        \hline
        
        Arterial Base Excess (BE) & 
        \cellcolor{green!10}numeric & meq/L & \\
        \hline
        
        Bicarbonate (HCO$_3$) & 
        \cellcolor{green!10}numeric & meq/L & \\
        \hline
        
        FiO$_2$ & 
        \cellcolor{green!10}numeric & fraction & \\
        \hline
        
        Glucose & 
        \cellcolor{green!10}numeric & mg/dL & \\
        \hline
        
        Blood Urea Nitrogen (BUN) & 
        \cellcolor{green!10}numeric & mg/dL & \\
        \hline
        
        Creatinine & 
        \cellcolor{green!10}numeric & mg/dL & \\
        \hline
        
        Magnesium (Mg$^{++}$) & 
        \cellcolor{green!10}numeric & mg/dL & \\
        \hline
        
        Serum Glutamic Oxaloacetic Transaminase (SGOT) & 
        \cellcolor{green!10}numeric & u/L & \\
        \hline
        
        Serum Glutamic Pyruvic Transaminase (SGPT) &
        \cellcolor{green!10}numeric & u/L & \\
        \hline
        
        Total Bilirubin (Total Bili) &
        \cellcolor{green!10}numeric & mg/dL & \\
        \hline
        
        White Blood Cell Count (WBC) &
        \cellcolor{green!10}numeric & E9/L & \\
        \hline
        
        Platelets Count (Platelets) &
        \cellcolor{green!10}numeric & E9/L & \\
        \hline
        
        PaO$_2$ &
        \cellcolor{green!10}numeric & mmHg & \\
        \hline
        
        Partial Pressure of CO$_2$ (PaCO$_2$) &
        \cellcolor{green!10}numeric & mmHg & \\
        \hline
        
        Lactate &
        \cellcolor{green!10}numeric & mmol/L & \\
        \hline
        
        Total Volume of Intravenous Fluids (Input Total) &
        \cellcolor{green!10}numeric & mL & \\
        \hline
        
        Intravenous Fluids of Each 4-Hour Period (Input 4H) &
        \cellcolor{green!10}numeric & mL & \\
        \hline
        
        Maximum Dose of Vasopressors in 4H (Max Vaso) &
        \cellcolor{green!10}numeric & mcg/kg/min & \\
        \hline
        
        Total Volume of Urine Output (Output Total) &
        \cellcolor{green!10}numeric & mL &\\
        \hline
        
        Urine Output in 4H (Output 4H) &
        \cellcolor{green!10}numeric & mL & \\
        \hline

    \end{tabular}
    
    \caption{\label{Tab:SepsisNumeric}Numeric Variables in the Sepsis Dataset\\
This table presents information related to the numeric variables used in the sepsis datasets; to be continued with the binary and categorical variables in Table \ref{Tab:SepsisNonNumeric}.
}
\end{table}

\newpage
\begingroup
\renewcommand{\arraystretch}{1.25}
\begin{table}[ht]
    \footnotesize
    \centering
    \begin{tabular}{|l||l|l|l|}
        \hline
        \textbf{Variable Name} & 
        \textbf{Data Type} & \textbf{Unit} &
        \textbf{Extra Notes}\\
        
        \hline
        \hline
        Gender &
        \cellcolor{cyan!10}binary & - - & {Male; \quad Female}\\
        \hline
        
        Readmission of Patient &
        \cellcolor{cyan!10}binary & - - & {False; \quad True}\\
        
        (Readmission) & & &\\
        \hline
        
        Mechanical Ventilation &
        \cellcolor{cyan!10}binary & - - & {False; \quad True}\\
        (Mech) & & &\\
        \hline
        \hline
        
        GCS &
        \cellcolor{brown!10}categorical & point & 
        {13 Classes}\\
        & & & {$3$; \quad$4$; \quad$5$; \quad$6$;}\\
        & & & {$7$; \quad$8$; \quad$9$; \quad$10$;}\\
        & & & {$11$; \quad$12$; \quad$13$;}\\
        & & & {$14$; \quad$15$}\\
        \hline
        
        Pulse Oximetry Saturation&
        \cellcolor{brown!10}categorical & $\%$ &
        {10 Classes (C)}\\
        (SpO$_2$) & & & {C1: $[0.00, 93.83)$; \quad C2: $[93.83, 95.14)$};\\
        & & & {C3: $[95.14, 96.00)$; \quad C4: $[96.00, 96.70)$};\\
        & & & {C5: $[96.70, 97.33)$; \quad C6: $[97.33, 98.00)$};\\
        & & & {C7: $[98.00, 98.60)$; \quad C8: $[98.60, 99.22)$};\\
        & & & {C9: $[99.22, 99.86)$; \quad C10: $[99.86, 100.0]$};\\
        \hline
        
        Temperature &
        \cellcolor{brown!10}categorical & Celsius &
        {10 Classes (C)}\\
        (Temp) & & & {C1: $[15.11, 35.95)$; \quad C2: $[35.95, 36.28)$};\\
        & & & {C3: $[36.28, 36.50)$; \quad C4: $[36.50, 36.69)$};\\
        & & & {C5: $[36.69, 36.88)$; \quad C6: $[36.88, 37.06)$};\\
        & & & {C7: $[37.06, 37.28)$; \quad C8: $[37.28, 37.56)$};\\
        & & & {C9: $[37.56, 37.93)$; \quad C10: $[37.93, 40.52]$};\\
        \hline
        
        Partial Thromboplastin Time &
        \cellcolor{brown!10}categorical & s & {10 Classes (C)}\\
        (PTT) & & & {C1: $[17.80, 24.53)$; \quad C2: $[24.53, 26.63)$};\\
        & & & {C3: $[26.63, 28.20)$; \quad C4: $[28.20, 29.60)$};\\
        & & & {C5: $[29.60, 31.45)$; \quad C6: $[31.45, 34.00)$};\\
        & & & {C7: $[34.00, 37.10)$; \quad C8: $[37.10, 42.80)$};\\
        & & & {C9: $[42.80, 57.90)$; \quad C10: $[57.90, 150.00]$};\\
        \hline
        
        Prothrombin Time &
        \cellcolor{brown!10}categorical & s & {10 Classes (C)}\\
        (PT) & & & {C1: $[9.90, 12.20)$; \quad C2: $[12.20, 12.90)$};\\
        & & & {C3: $[12.90, 13.30)$; \quad C4: $[13.30, 13.80)$};\\
        & & & {C5: $[13.80, 14.30)$; \quad C6: $[14.30, 14.90)$};\\
        & & & {C7: $[14.90, 15.90)$; \quad C8: $[15.90, 17.51)$};\\
        & & & {C9: $[17.51, 22.00)$; \quad C10: $[22.00, 146.70]$};\\
        \hline
        
        International Normalised Ratio &
        \cellcolor{brown!10}categorical & - - & {10 Classes (C)}\\
        (INR) & & & {C1: $[0.00, 1.00)$; \quad C2: $[1.00, 1.10)$};\\
        & & & {C3: $[1.10, 1.20)$}; \quad C4: $[2.21, 17.60]$\\
        & & & {C5: $[1.20, 1.30)$; \quad C6: $[1.30, 1.31)$};\\
        & & & {C7: $[1.31, 1.50)$; \quad C8: $[1.50, 1.70)$};\\
        & & & {C9: $[1.70, 2.21)$; \quad C10: $[2.21, 17.60]$};\\
        \hline

    \end{tabular}
    
    \caption{\label{Tab:SepsisNonNumeric}Non-Numeric Variables in the Sepsis Dataset\\
This table presents information related to the binary and categorical variables used in the sepsis dataset. It continues from Table \ref{Tab:SepsisNumeric}; and in addition, we listed the different levels available. 
}
\end{table}
\endgroup

\newpage

\newpage

\newpage

\newpage
\begin{table}[ht]
    \small
    \centering
    \begin{tabular}{|l||c|c|c|c|}
        \hline
        \textbf{Variable Name} & \textbf{\textcolor{white}{.}\hspace{3mm} KS Test \textcolor{white}{.}\hspace{3mm}} & 
        \textbf{\textcolor{white}{.}\hspace{3mm} t-Test \textcolor{white}{.}\hspace{3mm}} & 
        \textbf{\textcolor{white}{.}\hspace{3mm} F-Test \textcolor{white}{.}\hspace{3mm}} & 
        \textbf{Three Sigma Rule Test}\\
        \hline
        \hline
        
        Age & \cellcolor{cyan!10}$86/100$ & 
              \cellcolor{cyan!10}$93/100$ & 
              \cellcolor{magenta!10}$29/100$ & 
              \cellcolor{cyan!10}$100/100$\\
        \hline
        
        HR & \cellcolor{cyan!10}$91/100$ & 
              \cellcolor{cyan!10}$92/100$ & 
              \cellcolor{magenta!10}$53/100$ & 
              \cellcolor{cyan!10}$100/100$\\
        \hline
        
        Systolic BP & \cellcolor{cyan!10}$87/100$ & 
              \cellcolor{cyan!10}$92/100$ & 
              \cellcolor{magenta!10}$63/100$ & 
              \cellcolor{cyan!10}$100/100$\\
        \hline
        
        Mean BP & \cellcolor{cyan!10}$87/100$ & 
              \cellcolor{cyan!10}$90/100$ & 
              \cellcolor{magenta!10}$67/100$ & 
              \cellcolor{cyan!10}$100/100$\\
        \hline
        
        Diastolic BP & \cellcolor{magenta!10}$59/100$ & 
              \cellcolor{magenta!10}$64/100$ & 
              \cellcolor{cyan!10}$71/100$ & 
              \cellcolor{cyan!10}$100/100$\\
        \hline
        
        RR & \cellcolor{cyan!10}$82/100$ & 
              \cellcolor{cyan!10}$86/100$ & 
              \cellcolor{magenta!10}$57/100$ & 
              \cellcolor{cyan!10}$100/100$\\
        \hline
        
        K$^+$ & \cellcolor{magenta!10}$45/100$ & 
              \cellcolor{magenta!10}$63/100$ & 
              \cellcolor{magenta!10}$47/100$ & 
              \cellcolor{cyan!10}$100/100$\\
        \hline
        
        Na$^+$ & \cellcolor{cyan!10}$89/100$ & 
              \cellcolor{cyan!10}$90/100$ & 
              \cellcolor{magenta!10}$54/100$ & 
              \cellcolor{cyan!10}$100/100$\\
        \hline
        
         Cl$^-$ & \cellcolor{magenta!10}$57/100$ & 
              \cellcolor{magenta!10}$65/100$ & 
              \cellcolor{magenta!10}$40/100$ & 
              \cellcolor{cyan!10}$100/100$\\
        \hline
        
         Ca$^{++}$ & \cellcolor{cyan!10}$87/100$ & 
              \cellcolor{cyan!10}$86/100$ & 
              \cellcolor{cyan!10}$85/100$ & 
              \cellcolor{cyan!10}$100/100$\\
        \hline
        
        Ionised Ca$^{++}$& \cellcolor{magenta!10}$62/100$ & 
              \cellcolor{magenta!10}$53/100$ & 
              \cellcolor{magenta!10}$66/100$ & 
              \cellcolor{cyan!10}$100/100$\\
        \hline
        
        CO$_2$ & \cellcolor{cyan!10}$85/100$ & 
              \cellcolor{cyan!10}$92/100$ & 
              \cellcolor{magenta!10}$54/100$ & 
              \cellcolor{cyan!10}$100/100$\\
        \hline
        
        Albumin & \cellcolor{cyan!10}$93/100$ & 
              \cellcolor{cyan!10}$91/100$ & 
              \cellcolor{cyan!10}$82/100$ & 
              \cellcolor{cyan!10}$100/100$\\
        \hline
        
        Hb & \cellcolor{cyan!10}$82/100$ & 
              \cellcolor{cyan!10}$90/100$ & 
              \cellcolor{magenta!10}$47/100$ & 
              \cellcolor{cyan!10}$100/100$\\
        \hline
        
        pH & \cellcolor{cyan!10}$78/100$ & 
              \cellcolor{cyan!10}$78/100$ & 
              \cellcolor{cyan!10}$76/100$ & 
              \cellcolor{cyan!10}$100/100$\\
        \hline
        
        BE & \cellcolor{cyan!10}$91/100$ & 
              \cellcolor{cyan!10}$91/100$ & 
              \cellcolor{cyan!10}$80/100$ & 
              \cellcolor{cyan!10}$100/100$\\
        \hline
        
        HCO$_3$ & \cellcolor{cyan!10}$84/100$ & 
              \cellcolor{cyan!10}$91/100$ & 
              \cellcolor{magenta!10}$48/100$ & 
              \cellcolor{cyan!10}$100/100$\\
        \hline
        
        FiO$_2$ & \cellcolor{magenta!10}$29/100$ & 
              \cellcolor{cyan!10}$60/100$ & 
              \cellcolor{magenta!10}$61/100$ & 
              \cellcolor{cyan!10}$100/100$\\
        \hline
        
        Glucose & \cellcolor{cyan!10}$78/100$ & 
              \cellcolor{cyan!10}$86/100$ & 
              \cellcolor{magenta!10}$53/100$ & 
              \cellcolor{cyan!10}$100/100$\\
        \hline
        
        BUN & \cellcolor{cyan!10}$86/100$ & 
              \cellcolor{cyan!10}$92/100$ & 
              \cellcolor{cyan!10}$52/100$ & 
              \cellcolor{cyan!10}$100/100$\\
        \hline
        
        Creatinine & \cellcolor{magenta!10}$0/100$ & 
              \cellcolor{magenta!10}$5/100$ & 
              \cellcolor{magenta!10}$53/100$ & 
              \cellcolor{cyan!10}$88/100$\\
        \hline
        
        Mg$^{++}$ & \cellcolor{magenta!10}$12/100$ & 
              \cellcolor{magenta!10}$18/100$ & 
              \cellcolor{magenta!10}$50/100$ & 
              \cellcolor{cyan!10}$98/100$\\
        \hline
        
        SGOT & \cellcolor{magenta!10}$65/100$ & 
              \cellcolor{cyan!10}$72/100$ & 
              \cellcolor{cyan!10}$92/100$ & 
              \cellcolor{cyan!10}$97/100$\\
        \hline
        
        SGPT & \cellcolor{cyan!10}$96/100$ & 
              \cellcolor{cyan!10}$96/100$ & 
              \cellcolor{cyan!10}$91/100$ & 
              \cellcolor{cyan!10}$100/100$\\
        \hline
        
        Total Bili & \cellcolor{magenta!10}$52/100$ & 
              \cellcolor{cyan!10}$71/100$ & 
              \cellcolor{cyan!10}$89/100$ & 
              \cellcolor{cyan!10}$95/100$\\
        \hline
        
        WBC & \cellcolor{cyan!10}$83/100$ & 
              \cellcolor{cyan!10}$88/100$ & 
              \cellcolor{magenta!10}$44/100$ & 
              \cellcolor{cyan!10}$100/100$\\
        \hline
        
        Platelets & \cellcolor{magenta!10}$44/100$ & 
              \cellcolor{magenta!10}$58/100$ & 
              \cellcolor{magenta!10}$36/100$ & 
              \cellcolor{cyan!10}$100/100$\\
        \hline
        
        PaO$_2$ & \cellcolor{cyan!10}$79/100$ & 
              \cellcolor{cyan!10}$86/100$ & 
              \cellcolor{cyan!10}$93/100$ & 
              \cellcolor{cyan!10}$100/100$\\
        \hline
        
        PaCO$_2$ & \cellcolor{cyan!10}$99/100$ & 
              \cellcolor{cyan!10}$94/100$ & 
              \cellcolor{cyan!10}$78/100$ & 
              \cellcolor{cyan!10}$99/100$\\
        \hline
        
        Lactate & \cellcolor{magenta!10}$26/100$ & 
              \cellcolor{magenta!10}$28/100$ & 
              \cellcolor{cyan!10}$86/100$ & 
              \cellcolor{cyan!10}$100/100$\\
        \hline
        
        Input Total & \cellcolor{cyan!10}$83/100$ & 
              \cellcolor{cyan!10}$72/100$ & 
              \cellcolor{magenta!10}$6/100$ & 
              \cellcolor{cyan!10}$100/100$\\
        \hline
        
        Input 4H & \cellcolor{magenta!10}$69/100$ & 
              \cellcolor{cyan!10}$80/100$ & 
              \cellcolor{magenta!10}$14/100$ & 
              \cellcolor{cyan!10}$100/100$\\
        \hline
        
        Max Vaso & \cellcolor{magenta!10}$0/100$ & 
              \cellcolor{magenta!10}$6/100$ & 
              \cellcolor{cyan!10}$58/100$ & 
              \cellcolor{cyan!10}$80/100$\\
        \hline
        
        Output Total & \cellcolor{magenta!10}$66/100$ & 
              \cellcolor{magenta!10}$64/100$ & 
              \cellcolor{magenta!10}$13/100$ & 
              \cellcolor{cyan!10}$100/100$\\
        \hline
        
        Output 4H & \cellcolor{cyan!10}$80/100$ & 
              \cellcolor{magenta!10}$66/100$ & 
              \cellcolor{magenta!10}$13/100$ & 
              \cellcolor{cyan!10}$100/100$\\
        \hline
        \hline
        
        Gender & \cellcolor{cyan!10}$99/100$ & 
              \cellcolor{gray!10}- - & 
              \cellcolor{cyan!10}$96/100$ & 
              \cellcolor{gray!10}- -\\
        \hline
        
        Readmission & \cellcolor{cyan!10}$93/100$ & 
              \cellcolor{gray!10}- - & 
              \cellcolor{cyan!10}$75/100$ & 
              \cellcolor{gray!10}- -\\
        \hline
        
        Mech & \cellcolor{cyan!10}$98/100$ & 
              \cellcolor{gray!10}- - & 
              \cellcolor{cyan!10}$88/100$ & 
              \cellcolor{gray!10}- -\\
        \hline
        \hline
        
        GCS & \cellcolor{cyan!10}$98/100$ & 
              \cellcolor{gray!10}- - & 
              \cellcolor{cyan!10}$99/100$ & 
              \cellcolor{gray!10}- -\\
        \hline
        
        SpO$_2$ & \cellcolor{cyan!10}$97/100$ & 
              \cellcolor{gray!10}- - & 
              \cellcolor{cyan!10}$91/100$ & 
              \cellcolor{gray!10}- -\\
        \hline
        
        Temp & \cellcolor{cyan!10}$92/100$ & 
              \cellcolor{gray!10}- - & 
              \cellcolor{cyan!10}$97/100$ & 
              \cellcolor{gray!10}- -\\
        \hline
        
        PTT & \cellcolor{cyan!10}$96/100$ & 
              \cellcolor{gray!10}- - & 
              \cellcolor{cyan!10}$94/100$ & 
              \cellcolor{gray!10}- -\\
        \hline
        
        PT & \cellcolor{cyan!10}$97/100$ & 
              \cellcolor{gray!10}- - & 
              \cellcolor{cyan!10}$94/100$ & 
              \cellcolor{gray!10}- -\\
        \hline
        
        INR & \cellcolor{cyan!10}$99/100$ & 
              \cellcolor{gray!10}- - & 
              \cellcolor{cyan!10}$95/100$ & 
              \cellcolor{gray!10}- -\\
        \hline

    \end{tabular}
    
    \caption{\label{Tab:SepsisStats}Results on the hierarchical statistical tests for Sepsis.}
\end{table}

\newpage
\begin{table}[ht]
    \centering
    \begin{tabular}{|l||l|l|}
        \hline
        \textbf{Variable Name} & 
        \textbf{KL Divergence} from $\mathfrak{D}_\text{null}$ & 
        \textbf{KL Divergence} from $\mathfrak{D}_\text{alt}$ (Ours)\\
        \hline
        \hline
        Age                 & 0.2230 &      0.1945\\
        \hline

        HR                  & 0.0171 &      0.0978\\
        \hline

        Systolic BP               & 0.1438 &      0.0689\\
        \hline

        Mean BP              & 0.1520 &      0.0745\\
        \hline

        Diastolic BP               & 0.0179 &      0.1575\\
        \hline

        RR                  & 0.0789 &      0.0972\\
        \hline

        K$^{+}$                   & 0.0293 &      0.1905\\
        \hline

        Na$^{+}$                   & 0.0737 &      0.0511\\
        \hline

        Cl$^{-}$                   & 0.0618 &      0.1459\\
        \hline

        Ca$^{++}$                   & 0.1355 &      0.0288\\
        \hline

        Ionised Ca$^{++}$           & 0.1040 &      0.1382\\
        \hline

        CO$_{2}$                  & 0.1776 &      0.0336\\
        \hline

        Albumin             & 0.1259 &      0.0228\\
        \hline

        Hb                  & 0.0246 &      0.1049\\
        \hline

        pH                  & 0.0186 &      0.0942\\
        \hline

        BE                  & 0.0463 &      0.0456\\
        \hline

        HCO$_{3}$                 & 0.2082 &      0.0840\\
        \hline

        FiO$_{2}$                 & 0.5679 &      0.7077\\
        \hline

        Glucose             & 0.0179 &      0.0888\\
        \hline

        BUN                 & 0.2531 &      0.0675\\
        \hline

        Creatinine          & 0.0213 &      0.0088\\
        \hline

        Mg$^{++}$                   & 0.0241 &      0.3387\\
        \hline

        SGOT                & 0.0175 &      0.0237\\
        \hline

        SGPT                & 0.0302 &      0.0112\\
        \hline

        Total Bili           & 0.0276 &      0.0670\\
        \hline

        WBC            & 0.0249 &      0.0496\\
        \hline

        Platelets      & 0.1250 &      0.2407\\
        \hline

        PaO$_{2}$                 & 0.0793 &      0.0618\\
        \hline

        PaCO$_{2}$                & 0.0373 &      0.0419\\
        \hline

        Lactate             & 0.0593 &      0.2549\\
        \hline

        Input Total          & 0.0934 &      0.0962\\
        \hline

        Input 4H             & 0.3364 &      0.3682\\
        \hline

        Max Vaso             & 1.9127 &      2.1157\\
        \hline

        Output Total         & 0.1695 &      0.2147\\
        \hline

        Output 4H            & 0.3005 &      0.2988\\
        \hline

        Gender              & 0.0729 &      0.0019\\
        \hline

        Readmission                & 0.0080 &      0.0553\\
        \hline

        Mech                & 0.0108 &      0.0202\\
        \hline

        GCS                 & 0.0879 &      0.0168\\
        \hline

        SpO$_{2}$                 & 0.0322 &      0.0343\\
        \hline

        Temp                & 0.0735 &      0.0302\\
        \hline

        PTT                 & 0.0388 &      0.0219\\
        \hline

        PT                  & 0.0225 &      0.0642\\
        \hline

        INR                 & 0.0294 &      0.0745\\
        \hline

    \end{tabular}
    
    \caption{\label{Tab:SepsisKLD}A comparison of the KL divergence of the sepsis variables.}
\end{table}

\newpage

\newpage
\begin{table}[ht!]
    \centering
    \begin{tabular}{|l||l|l|r|}
        \hline
        \multicolumn{2}{|l|}{\textbf{Dataset}} & $U$ & \textbf{CAT}\\
        \hline
        \hline

        \multirow{2}{*}{Sepsis} &  $\mathfrak{D}_\text{null}$~\citep{kuo2022health} & \textbf{-2.729} & 100.00\%\\
        \cline{2-4}
         &  $\mathfrak{D}_\text{alt}$ (ours) & -2.559 & 100.00\%\\
        \hline
    \end{tabular}
    
    \caption{\label{Tab:LogUCover2}A Comparison of the log-cluster metric ($U$) and the category coverage (CAT).\\
    It is the lower the better ($\downarrow$) for $U$; and higher the better ($\uparrow$) for CAT.}
\end{table}

\newpage

\end{document}